\documentclass{article}
\usepackage{iclr2027_conference,times}

\usepackage{amsmath,amsfonts,bm}

\def\eqref#1{equation~\ref{#1}}

\def\1{\bm{1}}

\DeclareMathAlphabet{\mathsfit}{\encodingdefault}{\sfdefault}{m}{sl}
\SetMathAlphabet{\mathsfit}{bold}{\encodingdefault}{\sfdefault}{bx}{n}

\usepackage{hyperref}
\usepackage{url}
\usepackage{graphicx}
\usepackage{amsmath}
\usepackage{amssymb}
\usepackage{booktabs}
\usepackage{microtype}
\usepackage{subcaption}
\usepackage{placeins}
\usepackage{enumitem}
\usepackage{xcolor}
\usepackage{algorithm}
\usepackage{algpseudocode}

\newif\ifrevhl
\revhlfalse
\definecolor{linkblue}{RGB}{0,82,163}
\definecolor{revcolor}{RGB}{0,70,200}
\newcommand{\rev}[1]{\ifrevhl{\color{revcolor}#1}\else#1\fi}
\newenvironment{revblock}{\ifrevhl\color{revcolor}\fi}{}

\graphicspath{{figures/}}

\newcommand{\RPO}{RPO}
\newcommand{\clip}{\operatorname{clip}}
\newcommand{\sg}{\operatorname{sg}}

\title{Routing Policy Optimization for Online Diffusion Reinforcement Learning}

\author{%
Qiuhong Shen$^{1}$ \enspace Yiding Yang$^{2}$ \enspace Yuxin Zhang$^{2}$ \enspace
Xinwei Huang$^{2}$ \enspace Minxuan Lin$^{2}$ \enspace Bo Liu$^{2}$ \\
\bfseries Hao Kang$^{2}$ \enspace Haotian Yang$^{2}$ \enspace Yizhi Wang$^{2}$ \enspace
Chongyang Ma$^{2}$ \enspace Xinchao Wang$^{1}$\thanks{Corresponding author.} \\[4pt]
\normalfont $^{1}$National University of Singapore \quad $^{2}$ByteDance
}

\iclrfinalcopy

\begin{document}

\maketitle
\lhead{Preprint}

\begin{abstract}
Reinforcement learning can align visual generators with reward signals beyond their pretraining objectives, but existing score-function methods typically introduce stochasticity by perturbing the reverse denoising trajectory, leading to costly rollout-based optimization.
We instead ask whether stochastic exploration can be introduced inside the generator while preserving its original sampler.
\emph{Routing Policy Optimization} (\RPO) answers this question by formulating attention routing as a discrete policy.
At a selected denoising level, each query block stochastically attends to a subset of candidate key blocks, with Bernoulli probabilities derived from the generator's native query--key affinities.
The resulting routing masks define discrete actions that can be optimized with a score-function objective.
\RPO{} further uses a controlled two-rollout estimator: two rollouts share the same prompt, initial latent, routed denoising level, and routed layers, differing only in their sampled attention routes.
Their reward difference provides an advantage signal for the routing actions.
Because the routing policy reuses the generator's native attention parameters, the learned updates remain effective after routing is removed.
At inference, \RPO{} restores dense attention and retains the original deterministic sampler.
Experiments on text-to-image and text-to-video generation show that \RPO{} improves generation quality while \rev{using $4\times$ fewer rollouts per training step than Flow-GRPO-style methods and reaching its dense-model gain within 15 training steps}.
Please see our {\hypersetup{pdfborder={0 0 0}}\href{https://velox-world.com/rpo}{\textcolor{linkblue}{project page}}} for visual results.
\end{abstract}

\section{Introduction}
\label{sec:introduction}

Diffusion and flow models now generate high-quality images and videos from
natural-language prompts. DDPM establishes stochastic reverse denoising, while
DDIM exposes a deterministic sampling path under the same training
objective~\citep{ho2020ddpm,song2021ddim}. Flow matching and rectified flow
instead learn vector fields for continuous transport
~\citep{lipman2023flow,liu2023flow,sd3}. In parallel, latent diffusion,
Diffusion Transformers, and video diffusion models make high-resolution
visual generation tractable at scale~\citep{ldm,dit,cogvideox,wan2025}.
Learned preference models then expose alignment objectives beyond likelihood
and reconstruction~\citep{imagereward,pickscore,hpsv3}.

Online reinforcement learning can optimize these objectives, but its central
bottleneck is the policy. DDPO and DPOK treat stochastic denoising transitions
as actions~\citep{black2024ddpo,fan2023dpok}. Flow-GRPO and DanceGRPO extend
this construction to deterministic flow models by converting their ODE
samplers into stochastic SDEs~\citep{flowgrpo,dancegrpo}. Temporal windows,
shared branches, and one-step updates reduce the resulting rollout cost
~\citep{mixgrpo,tempflowgrpo,branchgrpo,flashgrpo}, while dense credit and
regulated clipping improve how terminal rewards shape sampler transitions
~\citep{densegrpo,star,spatialflowgrpo,grpoguard}. Forward-matching objectives
avoid reverse-trajectory likelihoods, but they also remove the explicit
sampled action~\citep{diffusionnft,dgpo,awm,pathspacerl,velocityrl}. What
remains missing is an explicit, tractable policy that preserves the original
deterministic sampler for both image and video generators.

Our key insight is that stochastic exploration need not perturb the denoising
trajectory. A diffusion transformer splits each latent into visual tokens, and
its attention layers already compute query-key affinities between them. If we
group neighboring tokens into blocks, each query block can attend to a sampled
subset of key blocks instead of all of them as dense attention does. We call
this choice of visible key blocks a \emph{route}. Sampling routes exposes a
discrete action without adding a policy head or modifying the sampler. We
introduce \emph{Routing Policy Optimization} (\RPO), which
samples an affinity-conditioned Bernoulli attention mask at one denoising
level. The conditional likelihood of the recorded mask history supplies a
score-function surrogate for the native attention parameters that produce the
route probabilities.

\begin{figure}[t]
\vspace{-2mm}
\centering
\includegraphics[width=0.98\linewidth]{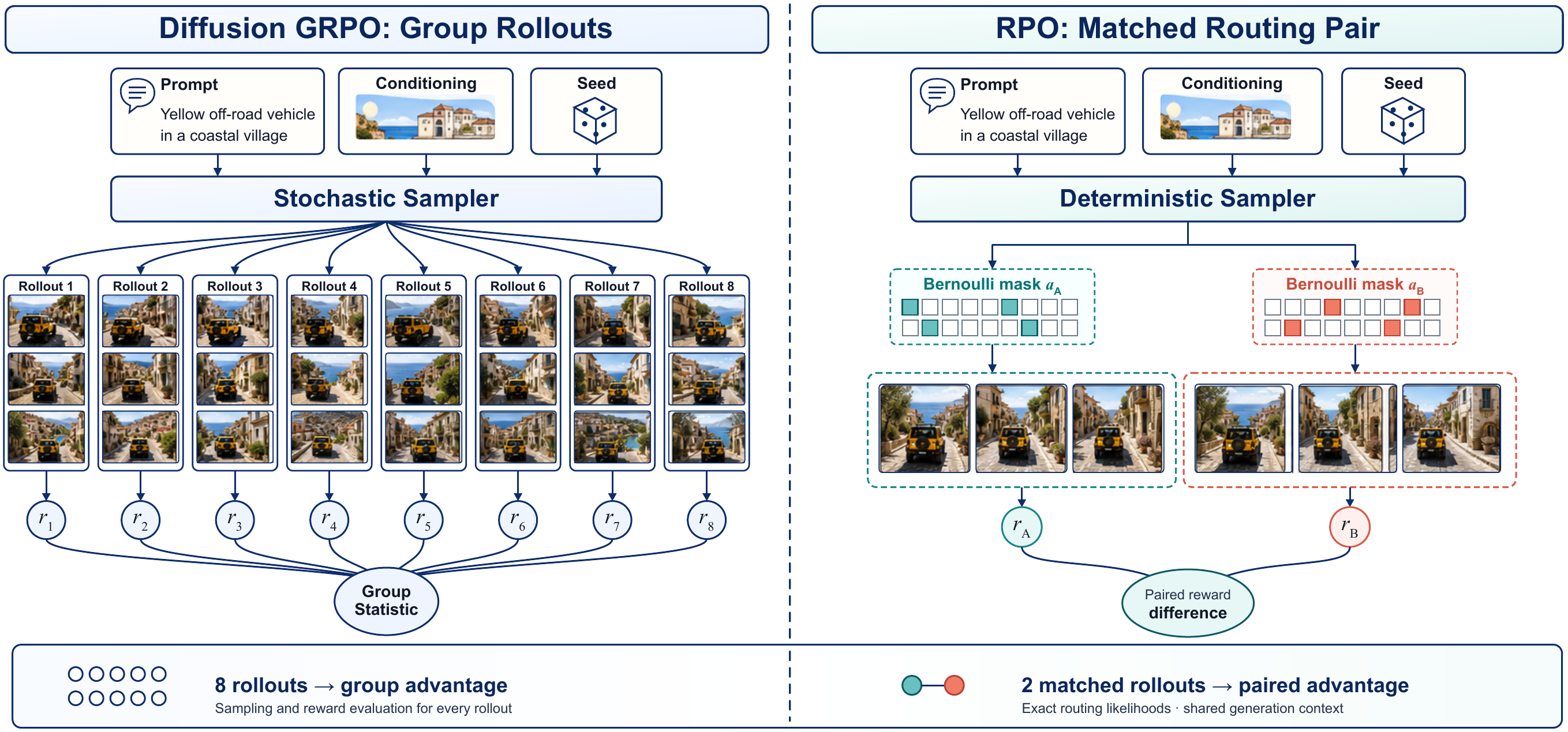}
\vspace{-2mm}
\caption{\textbf{A controlled routing pair replaces a large sampler-noise
group.} Sampler-space GRPO injects continuous noise into the reverse process
and estimates an advantage from many complete visual rollouts. \RPO{} retains
the original deterministic sampler and pairs two rollouts that share the
prompt, initial visual latent, routed denoising level, and routed layers. They
differ only in their Bernoulli attention masks, so their reward difference
measures the aggregate effect of the two recorded routing actions. Generated
visual outputs are schematic.}
\label{fig:motivation}
\end{figure}

The routing action also enables a controlled two-rollout estimator, as
Figure~\ref{fig:motivation} illustrates. Both members share every non-routing source of variation and differ only in their
sampled mask histories. Their signed reward difference therefore provides a
leave-one-out advantage for the two routing actions. During replay, later
probabilities condition on earlier recorded masks, which follows the
generator's forward execution order. A fixed logit offset sets the initial
sparsity operating point, while a nominal keep budget normalizes the loss.
Realized keep counts remain stochastic and can evolve with the attention
affinities.

Routing serves only as a training interface. Its likelihood updates native
query and key projections together with trainable upstream parameters that
shape later route scores. These parameters also participate in ordinary dense
attention. After training, we remove all masks and evaluate the updated
generator with dense attention and its original deterministic sampler. The
central empirical question is whether reward learned through temporary routing
transfers to the dense image or video generator used at inference.

We evaluate this question on Stable Diffusion 3.5 Medium and
Wan2.2-TI2V-5B~\citep{wan2025}, covering text-to-image and text-to-video
generation. Across both settings, \RPO{} improves reward-aligned generation
quality while retaining dense inference. On Wan2.2, the dense evaluation
gain reaches about \(+3\) by step 15 and remains near that level through step
50. \rev{Within this 50-step window, \RPO{} also attains the strongest overall
VBench score, and under one shared renderer Flow-GRPO needs 8 to 16 times more
rollout videos to reach the reward level of \RPO{}.} \rev{Each \RPO{} step generates 128 rollout videos, whereas each
sampler-space step generates 512. A full 50-step \RPO{} run takes a measured
\(25.4\) GPU-hours, compared with \(75.4\) for Flow-GRPO and \(102.3\) for
DanceGRPO to their evaluated checkpoints, although \RPO{} rolls out with a
slower backend (Appendix~\ref{app:cost}).} Our contributions can be summarized as \rev{follows}:

\begin{itemize}[leftmargin=1.4em,itemsep=1pt]
\item We introduce a sampler-preserving formulation for online alignment of
continuous visual generators. Its policy action is a training-time Bernoulli
mask over native attention affinities, with no auxiliary policy head or SDE
conversion.
\item We formulate executed routes as a conditional mask-history policy and
optimize them with a controlled two-rollout surrogate that holds the
non-routing generation context fixed.
\item We couple the routing policy to parameters reused by dense attention and
remove routing at inference, preserving the original image or video sampler.
\item We validate \RPO{} on Stable Diffusion 3.5 and Wan2.2, demonstrating
improved reward-aligned quality across image and video generation together
with \rev{a substantially smaller rollout budget} than sampler-space baselines.
\end{itemize}

\section{Related work}
\label{sec:related}

Reward alignment methods for visual generators differ mainly in where the
learning signal enters the generator. We organize prior work along this axis,
from the sampler trajectory to the attention computation.

\paragraph{Sampler-space policy gradients.}
DDPO and DPOK first treated stochastic denoising transitions as policy
actions~\citep{black2024ddpo,fan2023dpok}. Flow-GRPO and DanceGRPO carried
this view to flow matching and video by converting the deterministic sampler
into an SDE and estimating group-relative
advantages~\citep{flowgrpo,dancegrpo}. Follow-up work lowers the cost of these
stochastic trajectories. Temporal windows, shared-prefix branches, and
one-step updates reduce the number or cost of stochastic
transitions~\citep{mixgrpo,branchgrpo,flashgrpo}. TempFlow-GRPO concentrates
stochasticity at designated branch points for temporal credit
assignment~\citep{tempflowgrpo}, and TAGRPO aligns trajectories that start
from identical initial noise~\citep{tagrpo}. Dense temporal and spatial credit
or regulated clipping further improve how a terminal reward shapes each
transition~\citep{densegrpo,star,spatialflowgrpo,grpoguard}. These methods
keep the action inside the sampler, so exploration still perturbs the reverse
trajectory.

\paragraph{Objectives without sampler actions.}
A second family avoids sampler actions. Direct reward fine-tuning
backpropagates through differentiable image or video
rewards~\citep{draft,vader}. For black-box rewards, DiffusionNFT, DGPO, and
advantage-weighted matching replace reverse-trajectory likelihoods with
forward-process or preference objectives~\citep{diffusionnft,dgpo,awm}.
Adjoint matching, ELBO surrogates, path-space value gradients, and velocity
matching pursue the same goal through native regression
targets~\citep{ram,vgrpo,pathspacerl,velocityrl}. For few-step generators, a
hybrid scheme combines distribution matching, GRPO on stochastic intermediate
transitions, and reward backpropagation through the final
step~\citep{rtdm}. Offline preference optimization removes online rollouts
altogether, ranging from Diffusion-DPO to multidimensional and temporally dense
video preferences~\citep{wallace2024diffdpo,videodpo,videoalign,densedpo}.
These objectives leave inference untouched, but they give up an explicit
sampled action. In contrast, \RPO{} keeps a score-function policy with a
recorded action and places that action inside attention.

\paragraph{Discrete actions and attention routing.}
Discrete policies already exist in visual generation. MaskGRPO, Mask-GRPO,
and MaskFocus optimize token corruption, unmasking, or mask schedules in
natively discrete image generators~\citep{maskgrpomm,maskgrpot2i,maskfocus}.
Structured attention sparsity is also well studied, although mainly for
efficiency. Q/K-derived block gates and training-free compression expose
attention redundancy~\citep{seerattention,ditfastattn,sparsevideogen,sparsevdit}.
VSA, trainable dynamic routing, and hierarchical sparse attention learn such
sparsity inside image and video
generators~\citep{vsa,dyditpp,llsa,lightforcing,sparseforcing}.
Reinforcement learning has also selected sparse-attention levels or
recomputation schedules for acceleration~\citep{rapid3,recache}. KVPO comes
closest to alignment. It explores through stochastic historical-key routing in
autoregressive video, but it scores branches with a velocity-energy
surrogate~\citep{kvpo}. Other work optimizes attention allocation for
multimodal reasoning or routes preference credit through diffusion
attention~\citep{ral,topo}. Dropout-GRPO samples Bernoulli masks to diversify
deterministic latent reasoning~\citep{dropoutgrpo}. Our
paired estimator also builds on common random numbers and group-relative
baselines~\citep{asmussen2007ss,shao2024deepseekmath}, and 2-GRPO shows that
two language-model rollouts can retain most of a larger group's
performance~\citep{twogrpo}. \RPO{} combines these ingredients through a new
interface. Its action is a Bernoulli mask over native query-key affinities,
its likelihood updates parameters shared with dense attention, and the mask
disappears before inference.

\section{Preliminaries}
\label{sec:preliminaries}

\RPO{} fine-tunes a pretrained visual generator that already has a
deterministic sampler. This section fixes the notation for the sampler and the
reward, introduces attention over token blocks, and defines where \RPO{}
routes attention.

\paragraph{Denoising trajectory and reward.}
Let \(c\) denote a text prompt, and let \(N\) denote both the number of
denoising transitions and the largest noise level. Rollout \(i\) starts from
Gaussian noise and follows the deterministic sampler:
\begin{equation}
  \tau_i = (x_i^N, x_i^{N-1}, \ldots, x_i^0),
  \qquad x_i^N \sim \mathcal N(0, I),
  \qquad y_i = \mathrm{Dec}(x_i^0).
  \label{eq:trajectory}
\end{equation}
Here \(\tau_i\) is the reverse trajectory, and the noise level
\(t\in\{N,\ldots,0\}\) decreases during generation. The frozen decoder
\(\mathrm{Dec}\) maps the terminal latent \(x_i^0\) to the visual output
\(y_i\), which is an image or a video. A black-box reward model \(R_d\) then
scores the output on reward dimension \(d\), giving \(r_{i,d}=R_d(y_i,c)\).
Training uses these reward values but never differentiates through \(R_d\).

\paragraph{Attention over token blocks.}
\label{subsec:token-blocks}
The generator is a diffusion transformer that splits each latent into \(M\)
visual tokens, one per spatial patch of an image or spatiotemporal patch of a
video. In every self-attention layer and head, each token emits a query, a
key, and a value vector. The attention weights of a query are a softmax over
its scaled inner products with the keys, which we call query-key affinities.
Dense attention lets every query attend to every key. Block-sparse attention
coarsens this pattern. It arranges the tokens in a sequence, cuts the sequence
into \(Q\) contiguous \emph{token blocks} of \(B\) tokens, and applies a binary
block mask \(\mu\in\{0,1\}^{Q\times Q}\). A query token in block \(q\) attends
only to the key tokens of blocks \(j\) with \(\mu_{q,j}=1\), and its softmax
renormalizes over those keys. We order the tokens along a locality-preserving
Gilbert curve, so each block is a compact image or video patch. We call the set
of key blocks that query block \(q\) keeps its \emph{route}, and the choice of
routes \emph{routing}. Sparse-attention methods choose routes to save
computation. \RPO{} instead samples routes as its policy action during
training and restores dense attention at inference.

\paragraph{Routing sites and units.}
\label{subsec:routed-transition}
Routing acts at a single noise level for each prompt group:
\begin{equation}
  t^\star \sim \operatorname{Uniform}(\mathcal T_{\mathrm{route}}),
  \qquad
  \mathcal T_{\mathrm{route}}\subseteq\{1,\ldots,N\}.
  \label{eq:single-step-draw}
\end{equation}
The set \(\mathcal T_{\mathrm{route}}\) contains noise levels where a route
can still change the final output. All rollouts of a prompt group share
\(t^\star\). At \(t^\star\), a fixed subset of self-attention layers, the
\emph{routed layers}, apply block masks, while every other layer and noise
level runs dense attention. A \emph{routing unit} is one routing decision,
namely the route of query block \(q\) in routed layer \(\ell\) at level
\(t^\star\), written \(u=(t^\star,\ell,q)\). When each attention head \(h\)
routes independently, the unit becomes \(u=(t^\star,\ell,h,q)\). Each query block keeps a deterministic safety set
\(S_q\), which contains the block itself and a local spatial or spatiotemporal
neighborhood. The remaining blocks form the candidate set:
\begin{equation}
  C_q=\{1,\ldots,Q\}\setminus S_q.
  \label{eq:candidate-set}
\end{equation}
The route of every unit contains \(S_q\), and the policy decides only which
blocks of \(C_q\) it adds. Every query therefore keeps its local context, and
its softmax never runs over an empty set.
Appendix~\ref{app:gilbert-safety} details the traversal, the block sizes, and
the construction of \(S_q\). This setup confines all stochasticity to one routed
transition, while the prompt, the sampler, and the safety neighborhood stay
deterministic.

\vspace{-3mm}
\section{Routing Policy Optimization}
\label{sec:method}

\begin{figure}[t]
\vspace{-2mm}
\centering
\includegraphics[width=0.94\linewidth]{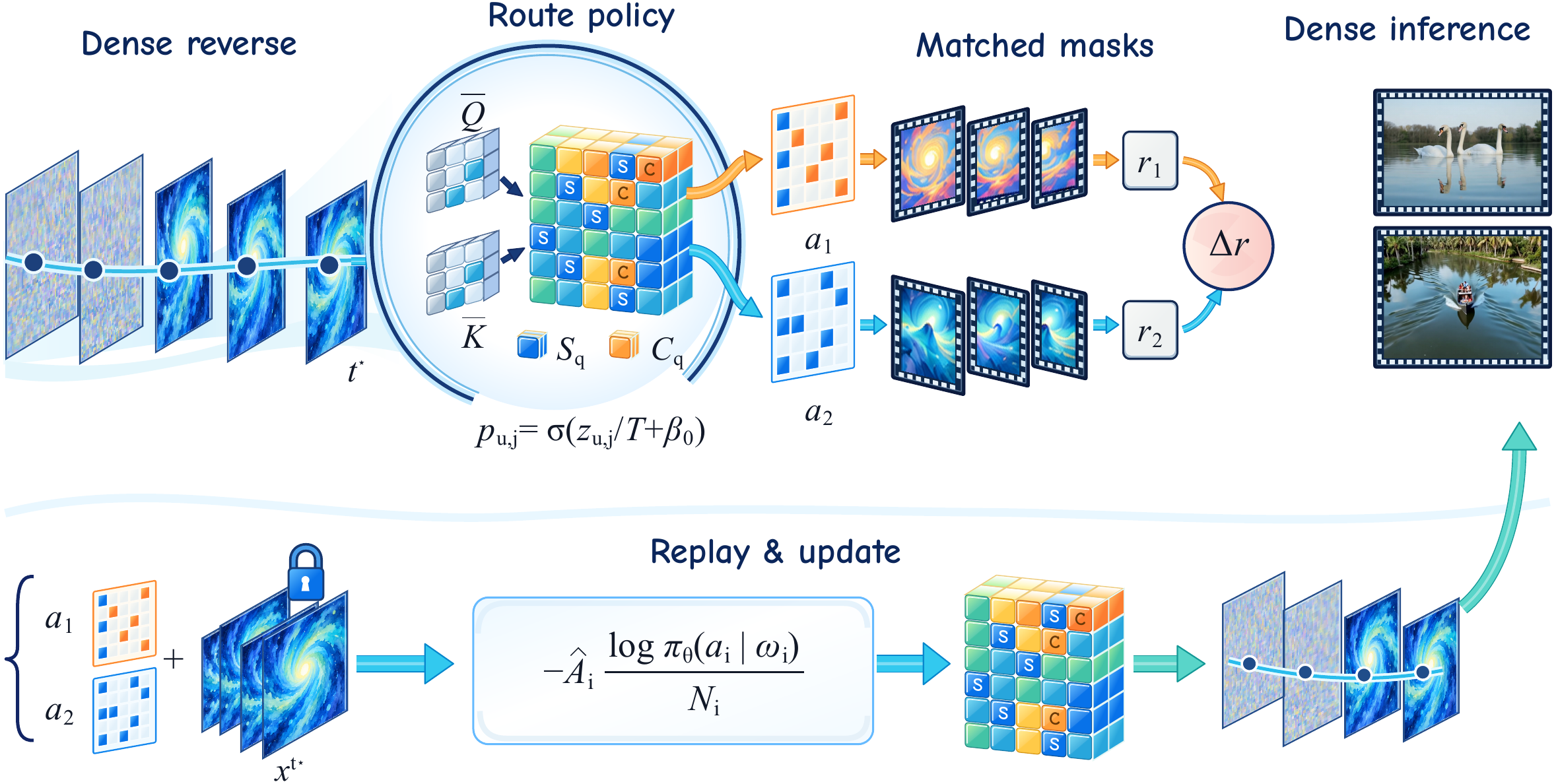}
\vspace{-2mm}
\caption{\textbf{A temporary route policy updates the dense generator.}
At the shared denoising level \(t^\star\), pooled native query-key affinities
define Bernoulli keep probabilities over candidate blocks \(C_q\), while safety
blocks \(S_q\) remain active. Two matched rollouts record independent mask
histories under the same prompt, content seed, routed level, and routed layers.
Their reward difference weights the conditional replay likelihood, whose
gradient updates attention parameters reused by the dense generator. At
inference, \RPO{} removes all masks and runs dense attention with the original
deterministic sampler.}
\label{fig:framework}
\end{figure}

Sampler-space methods explore by injecting noise into the reverse trajectory.
\RPO{} instead explores through a temporary attention mask at the routed level
\(t^\star\). We first define this mask policy from native query-key affinities
(Section~\ref{subsec:affinity-routing}). We then write the likelihood of a
recorded mask history (Section~\ref{subsec:conditional-replay}). Finally, we
pair two rollouts that differ only in their masks and turn their reward
difference into a policy-gradient update
(Section~\ref{subsec:paired-objective}). Figure~\ref{fig:framework} summarizes
this path from routed training to dense inference.

\vspace{-3mm}
\subsection{Affinity-conditioned Bernoulli routing}
\label{subsec:affinity-routing}

The central challenge is to expose a stochastic action without adding a
policy head or perturbing the sampler. A routed attention layer already
computes query-key affinities, so we reuse them as policy logits. For rollout
\(i\) and routing unit \(u\), let \(\bar Q_{i,u}(\theta)\in\mathbb
R^{d_{\mathrm h}}\) be the pooled query of block \(q\), namely the mean of its
token queries. Let \(\bar K_{i,j}(\theta)\in\mathbb R^{d_{\mathrm h}}\) be the
pooled key of candidate block \(j\in C_q\), defined in the same way. The policy keeps each candidate block independently:
\begin{equation}
\begin{aligned}
  z_{i,u,j}(\theta)
  &= \frac{\langle \bar Q_{i,u}(\theta),\bar K_{i,j}(\theta)\rangle}
         {\sqrt{d_{\mathrm h}}},
  &
  p_{i,u,j}(\theta)
  &= \sigma\!\left(\frac{z_{i,u,j}(\theta)}{T}+\beta_0\right),
  \\
  a_{i,u,j}
  &\sim\operatorname{Bernoulli}\!\left(p_{i,u,j}(\theta)\right),
  &
  j&\in C_q.
\end{aligned}
  \label{eq:bern-draw}
\end{equation}
Here \(\theta\) contains the trainable generator parameters, \(d_{\mathrm h}\)
is the attention-head width, \(T>0\) is a temperature, \(\sigma\) is the
logistic sigmoid, and \(\beta_0\) is a fixed logit offset. The sampled entry
\(a_{i,u,j}\in\{0,1\}\) keeps or drops block \(j\) for the query block of
unit \(u\). Under head-shared routing, we average the head-wise logits before
the sigmoid. Intuitively, blocks that the query already attends to strongly
are likely to stay visible, while the random draw still explores
alternatives. The offset sets the initial sparsity. It is neither learned nor
re-solved per row, so the number of kept blocks can change as the affinities
evolve.

A configured keep ratio \(\kappa\) defines a nominal budget for loss
normalization:
\begin{equation}
  K_u = \left\lceil \kappa |C_q| \right\rceil,
  \qquad
  N_i = \sum_{u\in\mathcal U_{t^\star}} K_u.
  \label{eq:nominal-budget}
\end{equation}
Here \(\mathcal U_{t^\star}\) is the set of routed units at the sampled noise
level, and \(K_u\) is the nominal selection count of unit \(u\). The
normalizer \(N_i\) depends only on the routed units, their candidate-set sizes,
and \(\kappa\), not on the
sampled keep count \(\sum_{j\in C_q}a_{i,u,j}\). As a result, a denser sample
cannot rescale its own loss. Native affinities thus decide which blocks stay
visible, while the mask cardinality remains genuinely stochastic.

\vspace{-3mm}
\subsection{Conditional mask-history likelihood}
\label{subsec:conditional-replay}

Independent draws within one unit do not yet give the likelihood of a
rollout. Routed units execute in a fixed forward order, and an earlier mask
changes the hidden states behind every later logit. We therefore score each
unit conditional on the recorded masks before it, denoted \(a_{i,<u}\):
\begin{equation}
  S_i(\theta)
  = \sum_{u\in\mathcal U_{t^\star}} \sum_{j\in C_q}
    \log\operatorname{Bern}\!\bigl(a_{i,u,j};\,
    p_{i,u,j}(\theta\mid a_{i,<u})\bigr)
  = \log \pi_\theta\!\left(a_i\mid \omega_i\right).
  \label{eq:rollout-total}
\end{equation}
Here \(\log\operatorname{Bern}(a;p)=a\log p+(1-a)\log(1-p)\), \(a_i\) is the
complete mask history of rollout \(i\), and \(\omega_i\) is its shared
context: the prompt, content seed, routed level, and routed layers. Replay
restarts from the stored latent \(x_i^{t^\star}\), executes the recorded masks
in order, and recomputes each probability. No mask is resampled. Therefore,
Equation~\ref{eq:rollout-total} is the exact likelihood of the recorded
history under the current policy. It is not the exact gradient of the full
generation and reward process, and Appendix~\ref{app:discussion} returns to
this gap. With conditional replay, the routing action gains a precise
log-likelihood that a policy gradient can weight.

\vspace{-3mm}
\subsection{Controlled two-rollout objective}
\label{subsec:paired-objective}
\label{subsec:advantage}

A score-function update is only as informative as its advantage. A reward
difference between unrelated rollouts mixes the effect of routing with changes
in content. We therefore generate each prompt group as a matched pair with
\(G=2\). Both members share the prompt \(c\) and a content seed that fixes the
initial latent and every non-routing source of randomness. Only their routing
seeds differ:
\begin{equation}
  \xi^1_{\mathrm{content}} = \xi^2_{\mathrm{content}},
  \qquad \xi^1_{\mathrm{route}} \neq \xi^2_{\mathrm{route}}.
  \label{eq:matched-seeds}
\end{equation}
Because the sampler is deterministic, the two members diverge only through
their mask histories. For member \(i\in\{1,2\}\), let \(\bar i\) denote the
other member of the pair. The partner reward then serves as a leave-one-out
baseline:
\begin{equation}
  A^{\mathrm{pair}}_{i,d} = r_{i,d}-r_{\bar i,d}.
  \label{eq:pair-advantage}
\end{equation}
Given the shared context, \(r_{\bar i,d}\) does not depend on member \(i\)'s
action, so it is a valid score-function baseline. The two advantages have
equal magnitude and opposite signs. Before replay, we winsorize the pair
differences of each reward dimension, divide them by a robust interquartile
scale, and sum the dimensions with weights \(w_d\). A stop-gradient turns the
result into a fixed replay weight \(\widehat A_i\), and the processing
preserves pair antisymmetry (Appendix~\ref{app:advantage}). The per-rollout
loss and the pair objective are:
\begin{equation}
  \mathcal L^{\mathrm{pg}}_i(\theta)
  = -\widehat A_i\frac{S_i(\theta)}{N_i},
  \qquad
  \mathcal L^{\mathrm{RPO}}(\theta)
  = \frac{1}{2}\sum_{i=1}^{2}\mathcal L^{\mathrm{pg}}_i(\theta).
  \label{eq:pg-loss}
\end{equation}
Here \(S_i\) is the history likelihood from Equation~\ref{eq:rollout-total},
and \(N_i\) is the action-independent normalizer from
Equation~\ref{eq:nominal-budget}. When both members route the same units with
the same candidate sets, \(N_1=N_2\). The update then raises the likelihood of the higher-reward mask
history and lowers that of the lower-reward one, in proportion to their
processed reward gap.

\paragraph{From routed training to dense inference.}
The route logits reuse native query and key projections and the upstream
parameters that shape them. The routing loss therefore updates parameters that
dense attention also uses. After training, we remove every mask and run the
updated image or video generator with dense attention and its original
deterministic sampler.

\vspace{-3mm}
\section{Experiments}
\label{sec:experiments}

Our experiments test whether reward learned through temporary routing transfers
to the dense generator used at inference. After the setup, we evaluate
text-to-image generation (Section~\ref{subsec:sd35}), compare with
sampler-space RL on video (Section~\ref{subsec:dancegrpo}), and measure VBench
transfer and training cost (Section~\ref{subsec:vbench}).

\vspace{-3mm}
\subsection{Setup}
\label{subsec:setup}

\paragraph{Image setting.}
We train Stable Diffusion 3.5 Medium~\citep{sd3} on prompts from the PickScore
SFW split~\citep{pickscore}. Each rollout generates a \(512\times512\) image
with a deterministic 10-step flow sampler and no classifier-free guidance. The
training reward is HPSv3~\citep{hpsv3} scaled by \(0.1\). After training, we
evaluate the dense model on DrawBench prompts~\citep{imagen}. An aesthetic
score and DeQA-Score~\citep{deqa} measure image quality, while
ImageReward~\citep{imagereward}, PickScore, and
UnifiedReward~\citep{unifiedreward} measure preference.

\paragraph{Video setting.}
We train Wan2.2-TI2V-5B~\citep{wan2025}, a text-to-video flow-matching generator, on a curated creative prompt
distribution. Each \rev{training} rollout renders 13 frames at \(704\times1280\) with
a fixed \rev{10-step Euler sampler and classifier-free guidance scale 5.
VBench and the qualitative samples use 50 sampling steps}. The reward is HPSv3 scaled by \(0.1\) and is
computed on uniformly sampled frames. During training, we probe the dense
model on a fixed 61-prompt validation set with a 10-step sampler.

\paragraph{Baselines and training.}
On Wan2.2, we compare against Flow-GRPO~\citep{flowgrpo} and
DanceGRPO~\citep{dancegrpo}. Both baselines train the same generator with the
same HPSv3 reward, but they explore by sampling SDE trajectories. DanceGRPO
uses groups of \(G=8\) and updates the full model. \RPO{} instead collects
64 matched pairs (\(G=2\)) per training step and routes a single denoising level.
\rev{Both baselines were launched with a 120-step schedule. We compare all
methods over the first 50 steps and report the full baseline runs in
Appendix~\ref{app:protocol}.}
Appendix~\ref{app:implementation} lists the routing and optimization
settings for both generators. Appendix~\ref{app:protocol} details the
evaluation protocols.

\vspace{-3mm}
\subsection{Text-to-image generation}
\label{subsec:sd35}

\begin{figure}[t]
\vspace{-2mm}
\centering
\includegraphics[width=0.86\linewidth]{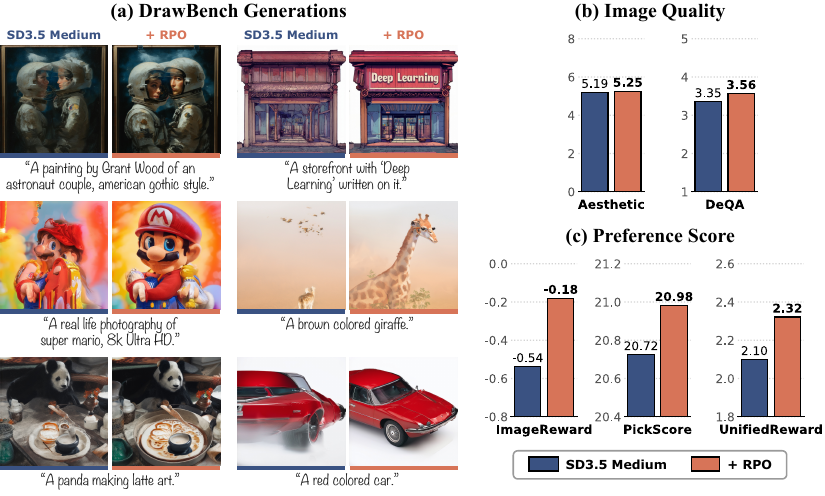}
\vspace{-2mm}
\caption{\textbf{\RPO{} on Stable Diffusion 3.5 Medium.}
(a) Samples before and after \RPO{} on DrawBench prompts. Each pair shares the
prompt and initial noise. (b, c) DrawBench scores over \(1{,}000\) images from
\(200\) prompts with \(5\) seeds, generated without classifier-free guidance.
\RPO{} improves desired property across all five reported metrics.}
\label{fig:sd35-drawbench}
\end{figure}

\begin{figure}[t]
\vspace{-2mm}
\centering
\begin{subfigure}[t]{0.45\linewidth}
\centering
\includegraphics[width=\linewidth]{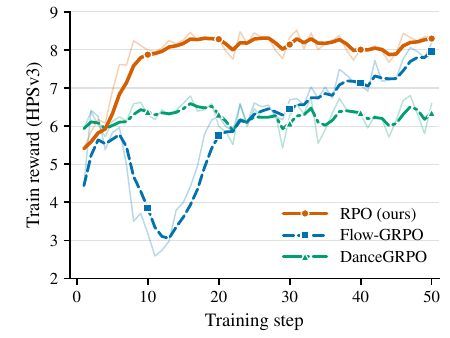}
\vspace{-3mm}
\caption{HPSv3 reward on the sampled training rollouts.}
\label{fig:dancegrpo-train}
\end{subfigure}
\hfill
\begin{subfigure}[t]{0.45\linewidth}
\centering
\includegraphics[width=\linewidth]{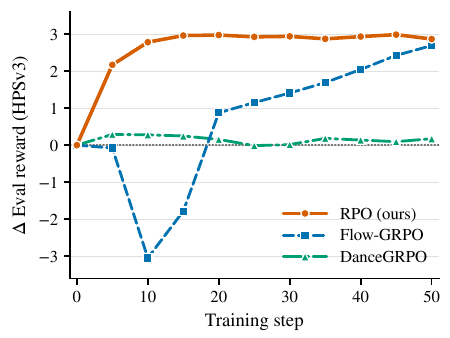}
\vspace{-3mm}
\caption{Evaluation reward relative to initial score.}
\label{fig:dancegrpo-eval}
\end{subfigure}
\vspace{-2mm}
\caption{\textbf{RPO improves reward rapidly and preserves its evaluation
gain.} We compare \RPO, Flow-GRPO, and DanceGRPO over the \rev{first} 50 training
steps on Wan2.2-TI2V-5B with HPSv3 reward. (a) Per-step reward on each method's
own training rollouts (faint) and its exponential moving average with factor
0.4 (bold). (b) Dense evaluation reward on the shared validation set,
relative to each method's initial score.}
\label{fig:dancegrpo-compare}
\end{figure}

Figure~\ref{fig:sd35-drawbench} first compares SD3.5 Medium before and after \RPO{}
training on DrawBench. \RPO{} improves all five metrics, although none of
them is the training reward. The preference metrics show the clearest gains.
ImageReward rises from \(-0.54\) to \(-0.18\), and UnifiedReward rises from
\(2.10\) to \(2.32\). Image quality also improves, with DeQA-Score rising from
\(3.35\) to \(3.56\). In the samples, the trained model writes the requested
storefront text and renders a clear brown giraffe, whereas the base model
misses both. Because each pair shares its prompt, initial noise, and sampler,
these gains come from the updated weights of the dense generator alone.

\vspace{-3mm}
\subsection{Comparison with sampler-space RL on video}
\label{subsec:dancegrpo}

Video generation raises the cost of every rollout, so learning speed matters
more. Figure~\ref{fig:dancegrpo-compare} tracks the first 50 training steps of
all three methods on Wan2.2. Figure~\ref{fig:dancegrpo-train} reports the
reward of the rollouts each method samples during training. \RPO{} raises this
reward from about 5.4 to a band around 8.2. DanceGRPO moves from about 5.9 to
a band around 6.2. Flow-GRPO first collapses and only returns to a comparable
level near step 50. However, training reward scores routed or SDE rollouts
rather than the dense model used at inference.

Figure~\ref{fig:dancegrpo-eval} therefore reports the \rev{dense model used at inference}. It
shows the dense evaluation reward as a gain over each method's own step-zero
score\rev{, because the step-zero scores of the three runs differ
(Appendix~\ref{app:protocol})}. \RPO{} gains
\(+2.17\) by step 5 and about \(+2.9\) by step 15, and it stays near \(+3\)
through step 50. Flow-GRPO follows its training reward. Its dense gain falls
to about \(-3\) at step 10 and then recovers gradually to \(+2.7\) at step 50.
DanceGRPO peaks at \(+0.29\) at step 5 and then stays between about \(0\) and
\(+0.2\). \RPO{} thus reaches its final dense gain within 15 steps and keeps
it. In contrast, Flow-GRPO needs the whole window to approach that level, and
DanceGRPO accumulates no gain. \rev{Because the step-zero scores differ, we
also re-render saved checkpoints of the three runs with one shared renderer
and identical noise, so every model starts from the same score of \(5.29\)
(Table~\ref{tab:shared-renderer}). \RPO{} reaches \(7.87\) after \(2{,}560\)
rollout videos. Flow-GRPO scores \(6.63\) after \(20{,}480\) videos and first
exceeds the \RPO{} level at its next saved checkpoint, after \(40{,}960\).
DanceGRPO stays at \(5.32\) after \(25{,}600\). With a larger budget,
Flow-GRPO then rises above \RPO{}. Our efficiency claim therefore concerns the
rollout budget needed to reach a given reward, not the final reward under an
unlimited budget.}

\vspace{-3mm}
\subsection{VBench transfer and training cost}
\label{subsec:vbench}

A single reward scalar compresses appearance and semantics, so reward gains
alone do not establish broad video quality. We therefore evaluate the base
model and the three trained checkpoints on VBench, which decomposes video
generation into 16 fine-grained dimensions with tailored prompts and
evaluation procedures~\citep{vbench}. Each model generates \(6{,}220\)
13-frame videos from shared prompts and initial noise with a deterministic
sampler (Appendix~\ref{app:protocol}).

\begin{table}[t]
\centering
\caption{\textbf{Comparison on VBench across multiple dimensions with
DanceGRPO and Flow-GRPO, plus overall score and training cost.} Scores are
percentages. The Wan2.2 rows share one generation configuration. CogVideoX-5B and
HunyuanVideo use the same prompts, frame count, denoising steps, and scoring,
but keep their native resolution and guidance. \rev{Training GPU-hours count
the H100 cost of reaching each evaluated checkpoint: step 50 for \RPO{} and
DanceGRPO, and step 40 for Flow-GRPO, its closest saved checkpoint. All costs
are measured. \RPO{} rolls out with a batched diffusers backend and the
baselines with vLLM-Omni (Appendix~\ref{app:cost}). All 16 dimensions are in
Table~\ref{tab:vbench-full}.}}
\label{tab:vbench}
\vspace{-2mm}
\begingroup
\scriptsize
\setlength{\tabcolsep}{2.4pt}
\setlength{\fboxsep}{1.2pt}
\renewcommand{\arraystretch}{1.12}
\newcommand{\best}[1]{\colorbox{blue!10}{\strut\textbf{#1}}}
\resizebox{\linewidth}{!}{%
\begin{tabular}{@{}lrrrrrrrrrr@{}}
\toprule
\textbf{Method} & \shortstack{\textbf{Train}\\\textbf{GPU-h}} &
\shortstack{\textbf{Aesthetic}\\\textbf{quality}} &
\shortstack{\textbf{Imaging}\\\textbf{quality}} &
\shortstack{\textbf{Subject}\\\textbf{consistency}} &
\shortstack{\textbf{Motion}\\\textbf{smoothness}} &
\shortstack{\textbf{Object}\\\textbf{class}} &
\shortstack{\textbf{Multiple}\\\textbf{objects}} &
\shortstack{\textbf{Human}\\\textbf{action}} & \textbf{Scene} &
\shortstack{\textbf{Overall}\\\textbf{score}} \\
\midrule
CogVideoX-5B & -- & 60.17 & 61.92 & 96.50 & 97.42 & 82.44 & 57.29 & 86.40 & 42.73 & 81.30 \\
HunyuanVideo & -- & 63.05 & 63.78 & 98.76 & 99.42 & 65.79 & 56.46 & 83.20 & 27.01 & 78.69 \\
\midrule
Wan2.2-TI2V-5B & -- & 61.91 & 67.71 & 98.37 & 98.76 & 81.01 & 59.15 & 79.40 & 37.41 & 81.61 \\
\textbf{\RPO{}} & \rev{25.4} & \best{62.64} & \best{68.65} & \best{98.69} & \best{99.06} & \best{85.41} & \best{61.30} & \best{84.20} & \best{40.84} & \best{82.00} \\
DanceGRPO & 102.3 & 61.74 & 67.69 & 98.38 & 98.75 & 80.81 & 59.34 & 80.00 & 36.95 & 81.46 \\
Flow-GRPO & 75.4 & 60.09 & 67.43 & 97.59 & 97.88 & 79.10 & 59.98 & 81.80 & 36.73 & 81.19 \\
\bottomrule
\end{tabular}}
\endgroup
\end{table}

Table~\ref{tab:vbench} lists \rev{8 of the 9} dimensions where \RPO{} exceeds both RL
baselines, together with the overall score and training cost. \rev{The ninth,
color, is listed in Table~\ref{tab:vbench-full}, because \RPO{} stays below the
base model there. Across all
16 dimensions (Table~\ref{tab:vbench-full}), \RPO{} exceeds the base model on
10 dimensions, and it exceeds DanceGRPO on 11 and Flow-GRPO on 11.} The largest
gains over the base model are semantic. Object class improves by \(+4.40\),
human action by \(+4.80\), scene by \(+3.43\), and multiple objects by
\(+2.15\). \RPO{} also improves aesthetic quality, imaging quality, subject
consistency, and motion smoothness over the base model. \rev{Among the
checkpoints in Table~\ref{tab:vbench}, it is the only one above the base model
overall, and a second training seed scores \(82.24\)
(Table~\ref{tab:vbench-extra}).} \rev{Its training cost is also
structurally smaller. Each step generates 128 rather than 512 rollout videos
and replays one rather than two denoising transitions per video. Its measured
cost of \(25.4\) GPU-hours is \(3.0\times\) and \(4.0\times\) below the
measured costs of the evaluated Flow-GRPO and DanceGRPO checkpoints, even
though its rollouts run on a slower backend (Appendix~\ref{app:cost}).}

The remaining dimensions reveal trade-offs. \RPO{} lowers color and appearance
style relative to the base model, although it still leads the trained
baselines on color. Flow-GRPO raises dynamic degree by \(12.50\) points but
loses consistency, smoothness, aesthetic quality, color, and object-class
accuracy. \RPO{} therefore transfers reward mainly through semantic alignment,
whereas Flow-GRPO trades visual quality for motion. \rev{The step-40
Flow-GRPO checkpoint already has a dense HPSv3 gain of \(+2.04\), yet it falls
below the base model on 11 of the 16 VBench dimensions. A higher HPSv3 score
therefore does not by itself indicate better videos. Flow-GRPO matches the
\RPO{} total only at steps 80 and 120 (\(82.02\) and \(81.99\)), after
\(6.4\) and \(9.6\) times as many rollout videos as \RPO{}, while its dynamic degree
falls to \(33.89\), below the base model.} The four Wan2.2 overall
scores span only \(0.81\) points on one seed set, so we read the aggregate
ranking as directional. \rev{A paired bootstrap over prompts gives \(+0.39\)
\([-0.01,+0.79]\) for the total gain of \RPO{} and \(+0.43\) \([+0.02,+0.85]\)
for its quality score (Table~\ref{tab:bootstrap}). The \RPO{} checkpoints at
steps 20 and 50 score \(82.03\) and \(82.00\), so the VBench gain is stable
over training.} CogVideoX-5B~\citep{cogvideox} and
HunyuanVideo~\citep{hunyuanvideo} serve as external references under matched
prompts and scoring.

\begin{figure}[t]
\centering
\includegraphics[width=0.73\linewidth]{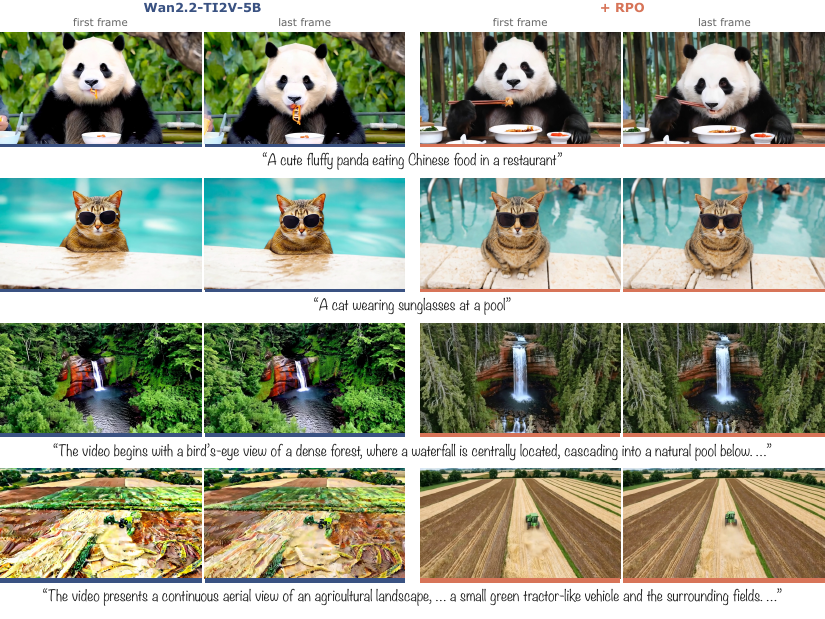}
\vspace{-2mm}
\caption{\textbf{Text-to-video samples from Wan2.2-TI2V-5B before and after
RPO.} Each row shows one prompt. For each model we show the first and last of
the 13 generated frames (\(1280\times704\), 50 sampling steps). Both models
use the same prompt and initial noise. RPO follows the prompt more closely and
replaces distorted, painterly textures with coherent, realistic scenes.}
\label{fig:qualitative}
\end{figure}

Figure~\ref{fig:qualitative} makes these semantic gains visible on four
held-out prompts, where the \RPO{} videos follow the prompts more closely.
Both checkpoints run dense attention with the same prompt, initial noise, and
sampler, so they differ only in their weights. Across image and video, the gains learned through routing
persist after the masks are removed.

\vspace{-3mm}
\section{Discussion and limitations}
\label{sec:discussion}

\RPO{} places its policy action inside attention, and this design choice also
defines its current limits. First, the framework is confined to attention.
Its action is a mask over key-value blocks, so it requires a backbone that
computes query-key affinities. We have not yet generalized routing to other
architectures, where no such affinity exists to reuse as a policy logit.
Second, the routing score is coarse. Each route logit is the inner product of
a pooled query and a pooled key, and pooling averages away token-level
structure within a block. The resulting score can therefore be noisy, and a
more fine-grained score design could give the policy a more faithful view of
which blocks matter. Finally, \RPO{} currently adapts to an existing
generator rather than shaping it. Co-designing the network architecture with
\RPO{} is a promising direction for achieving both higher efficiency and
larger RL gains.

\begin{revblock}
Further limits concern the scope of our evidence. Our efficiency claim
concerns the rollout budget needed to reach a given reward. With a much
larger budget, Flow-GRPO reaches a higher HPSv3 score and matches the VBench
total of \RPO{} (Appendix~\ref{app:ablations}), and we do not claim that \RPO{}
attains a higher final reward. The reported Wan2.2 run is the better of two
training seeds (\(+2.86\) and \(+2.40\)). Our single-seed ablations find that
neither the specific affinity pattern nor matched content seeds are required
for the gain, whereas the ratio-clipped second minibatch is. In addition, every video experiment uses 13-frame clips
for all methods, because rollout cost dominates training. Longer clips remain
to be evaluated.
\end{revblock}

\vspace{-3mm}
\section{Conclusion}
\label{sec:conclusion}

In this work, we addressed the cost of online reinforcement learning for
visual generators without perturbing their reverse denoising trajectories.
We introduced Routing Policy Optimization, a sampler-preserving method that
draws block-wise Bernoulli routes from native query-key affinities at one
selected denoising level. A controlled pair holds the non-routing generation
context fixed, so its reward difference weights the conditional likelihood of
the two recorded mask histories. Because the routing policy updates native
attention parameters, inference removes the masks and restores dense attention
under the original deterministic sampler. Experiments on Stable Diffusion 3.5
and Wan2.2 demonstrate improved reward-aligned quality across text-to-image and
text-to-video generation while requiring \rev{a substantially smaller rollout
budget} than sampler-space baselines. By moving stochastic exploration from the
sampler into attention, \RPO{} provides an efficient policy interface for
reward-aligned visual generation without altering the inference process.

\subsection*{AI use statement}

We used generative AI tools to polish the writing of this paper, namely to
edit the text for grammar, readability, and concision. We did not use
generative AI tools for any task that requires disclosure under the ICLR 2027
AI policy. In particular, AI tools did not propose hypotheses, develop the
method or its derivations, design experiments, implement code, generate or
clean data, or interpret results. Writing proofs, translation, and qualitative
data analysis are not applicable to this work. The authors reviewed every
AI-assisted edit to confirm that it preserves the intended technical meaning.
We take responsibility for the final content of this work, including all text,
claims, and results.

\subsection*{Ethics statement}

Visual generative models can amplify reward-model biases and hallucinate
plausible content, and \RPO{} is a general policy-gradient tool. We use a
public preference reward model on synthetic prompts and do not fine-tune on
user data. Reward hacking is a known risk of online reinforcement learning
for generation and is discussed in Appendix~\ref{app:discussion}.

\subsection*{Reproducibility statement}

Sections~\ref{sec:preliminaries} and~\ref{sec:method} state the routing policy
and the estimator in full. Section~\ref{subsec:setup} and
Appendix~\ref{app:implementation} record the models, prompt sets, reward model,
group size, keep ratio, temperature, offset, and update budget for each
generator. Appendix~\ref{app:protocol} describes every evaluation protocol. The
supplementary material includes a code release that reproduces the training
loop.


\appendix

\section{Gilbert blocks and the safety set}
\label{app:gilbert-safety}

Section~\ref{subsec:routed-transition} places every routing decision on a block
of visual tokens and protects a local safety set \(S_q\). It states both
constructions briefly and assumes a single block size \(B\). This appendix
gives the full construction that both generators use. We first explain why the
token order matters and define the frame-sliced Gilbert order. We then
decouple the block size into query, key, and safety partitions. Finally, we
define \(S_q\) and the executed attention mask, and we report the resulting
block and candidate counts for Wan2.2 and Stable Diffusion 3.5.

\paragraph{Why the token order matters.}
A routing unit pools a contiguous run of tokens into one query or key vector,
and one Bernoulli draw keeps or drops a whole run. Both generators store
visual tokens in raster order, which scans each latent frame row by row. A
contiguous raster run is therefore a thin horizontal strip. On the
\(32\times32\) token grid of Stable Diffusion 3.5, a 32-token run is one entire
image row. Its pooled key averages tokens that lie far apart, while its
vertical neighbors lie a full row away in the sequence. As a result, a route
decision on a raster run couples unrelated content, and a small spatial
neighborhood cannot be expressed as a few runs. The token order should instead
turn every contiguous run into a compact spatial patch.

\begin{figure}[t]
\centering
\includegraphics[width=\linewidth]{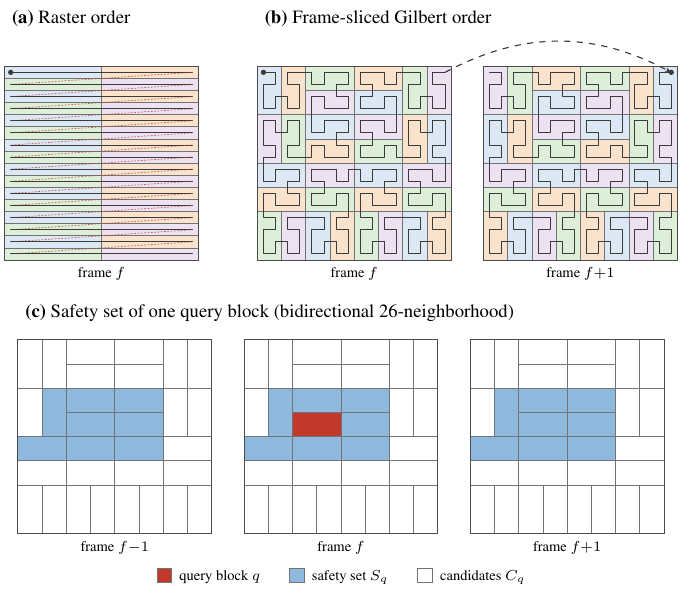}
\caption{\textbf{Gilbert blocks and the safety set on toy \(16\times16\)
frames with 8-token blocks.} (a)~Raster order turns each block into a
horizontal strip and jumps back at every row end (red dotted segments).
(b)~The frame-sliced Gilbert order turns each block into a compact
\(2\times4\) or \(4\times2\) patch. The curve continues from the end of frame
\(f\) to the same position in frame \(f{+}1\) (dashed arrow) and traverses that
frame in a reflected direction. (c)~The safety set \(S_q\) of one query block
\(q\) when \(B_{\mathrm q}=B_{\mathrm k}=B_{\mathrm s}=8\). Because attention
is bidirectional, \(S_q\) extends to both the previous and the next frame. All
remaining blocks form the candidate set \(C_q\).}
\label{fig:app-gilbert-safety}
\end{figure}

\subsection{Frame-sliced Gilbert order}
\label{app:gilbert-order}

A generalized Hilbert curve, or Gilbert curve, visits every cell of a
rectangle of arbitrary size exactly once and moves to an adjacent cell at every
step. Jenga~\citep{jenga} partitions video tokens along such a curve for
block-sparse attention, and we adopt the frame-sliced variant of its released
implementation. Let the visual latent contain \(F\) frames of
\(H\times W\) tokens, so it holds \(M=FHW\) tokens. Token \(p=(f,m,n)\) lies
in frame \(f\in\{0,\ldots,F-1\}\), row \(m\), and column \(n\). The order
assigns each token a rank:
\begin{equation}
  \gamma(f,m,n) = fHW + g_{H,W}\!\left(\phi_f(m,n)\right).
  \label{eq:gilbert-rank}
\end{equation}
Here \(g_{H,W}\) assigns the ranks \(1,\ldots,HW\) along the two-dimensional
Gilbert curve of an \(H\times W\) grid, which starts and ends at the two
corners of one longer side. The map \(\phi_f\) reflects the rows, the columns,
or both, so that frame \(f\) starts at the corner nearest to the end of frame
\(f-1\). The first map \(\phi_0\) is the identity, and the offset \(fHW\)
keeps every frame contiguous. Intuitively, the curve fills one frame, steps to
the same spatial position in the next frame, and fills that frame in a
reflected direction. Every step therefore moves to a neighboring token,
including the step across a frame boundary. Figure~\ref{fig:app-gilbert-safety}
contrasts this order with the raster order.

Contiguous runs of this order are compact patches. On the \(32\times32\) grid
of Stable Diffusion 3.5, every 32-token run is a \(4\times8\) or \(8\times4\)
rectangle, and every 128-token run is a \(16\times8\) or \(8\times16\)
rectangle. On the \(22\times40\) frames of Wan2.2, every 16-token run stays
inside one frame, and the longer side of its bounding box spans 5.7 tokens on
average and 8 tokens at most. The same run in raster order would be a
\(1\times16\) strip.

\paragraph{Bidirectional attention.}
Both generators denoise all visual tokens jointly with bidirectional
self-attention, so every query can attend to every frame. The order therefore
depends only on the latent grid and not on a generation schedule. We compute it
once per resolution and reuse it for all routed layers, heads, noise levels,
and rollouts.

\paragraph{Reordering inside attention.}
The order acts only inside each routed self-attention layer. After the native
projections, query-key normalization, and rotary position embedding where the
generator uses one, we permute the queries, keys, and values into Gilbert
order. We then apply the routed mask and permute the output back to raster
order. Dense attention is equivariant to a shared permutation of its queries,
keys, and values, and every positional signal is attached before the
permutation. The reordering thus changes which token pairs share a routing
decision, but it leaves positions and the dense computation unchanged. At
inference, \RPO{} runs the stock dense attention without any reordering.

\subsection{Query, key, and safety blocks}
\label{app:three-blocks}

A single block size ties three different roles together. The query block sets
the number of routing units, the key block sets the resolution of the action,
and the safety block sets the extent of the protected neighborhood. The
implementation therefore partitions the Gilbert sequence at three independent
sizes:
\begin{equation}
  \mathcal Q_q=\{p:\lceil\gamma(p)/B_{\mathrm q}\rceil=q\},
  \quad
  \mathcal K_j=\{p:\lceil\gamma(p)/B_{\mathrm k}\rceil=j\},
  \quad
  \mathcal R_b=\{p:\lceil\gamma(p)/B_{\mathrm s}\rceil=b\}.
  \label{eq:three-partitions}
\end{equation}
Here \(\mathcal Q_q\) is query block \(q\in\{1,\ldots,Q\}\) with
\(Q=\lceil M/B_{\mathrm q}\rceil\), \(\mathcal K_j\) is key block
\(j\in\{1,\ldots,J\}\) with \(J=\lceil M/B_{\mathrm k}\rceil\), and
\(\mathcal R_b\) is safety block \(b\). The sizes \(B_{\mathrm q}\),
\(B_{\mathrm k}\), and \(B_{\mathrm s}\) count Gilbert-ordered tokens, and the
last block of each partition may be shorter. For routing unit
\(u=(t^\star,\ell,h,q)\), the pooled query \(\bar Q_{i,u}\) in
Equation~\ref{eq:bern-draw} is the mean of the head-\(h\) queries over
\(\mathcal Q_q\), and the pooled key \(\bar K_{i,j}\) is the mean of the
head-\(h\) keys over \(\mathcal K_j\). One Bernoulli entry \(a_{i,u,j}\) then
keeps or drops key block \(\mathcal K_j\) for every query in \(\mathcal Q_q\).
A smaller \(B_{\mathrm k}\) yields more candidates and finer masks, whereas
\(B_{\mathrm s}\) alone controls how much local context stays protected. When
\(B_{\mathrm q}=B_{\mathrm k}=B_{\mathrm s}=B\), the three partitions coincide,
\(J=Q\), and the construction reduces to Section~\ref{subsec:routed-transition}.
This decoupling refines the routing action without shrinking the protected
neighborhood.

\subsection{Safety set and executed mask}
\label{app:safety-set}

Bernoulli routing makes the keep count random, and a single draw may drop
every candidate of a unit. Without protection, a query could lose all of its
keys, which leaves the attention softmax undefined, or lose its local context
at random. The safety set removes both failure modes by construction. Two
tokens \(p=(f,m,n)\) and \(p'=(f',m',n')\) are neighbors when
\(\max(|f-f'|,|m-m'|,|n-n'|)\le1\). This is the 26-neighborhood in the
spatiotemporal grid, and it reduces to the 8-neighborhood for images, where
\(F=1\). Following the block adjacency mask of Jenga~\citep{jenga}, safety
blocks \(b\) and \(b'\) are adjacent, written \(b\approx b'\), when some
token of \(\mathcal R_b\) neighbors some token of \(\mathcal R_{b'}\). Every
safety block is adjacent to itself. Let \(s(p)\) denote the
safety block that contains token \(p\). We define:
\begin{equation}
  \mathcal E_{\mathrm{safe}}
  = \{(p,p') : s(p)\approx s(p')\},
  \quad
  S_q = \{j : \mathcal Q_q\times\mathcal K_j\subseteq
  \mathcal E_{\mathrm{safe}}\},
  \quad
  C_q = \{1,\ldots,J\}\setminus S_q .
  \label{eq:safety-set-full}
\end{equation}
Here \(\mathcal E_{\mathrm{safe}}\) is the set of query-key token pairs that
are always attended. A key block enters \(S_q\) only when every pair between it
and query block \(q\) is safe. The candidate set \(C_q\) collects all other key
blocks, which matches Equation~\ref{eq:candidate-set} with \(Q\) replaced by
\(J\). Bernoulli entries exist only for \(j\in C_q\), so safety blocks carry no
probability and add no term to \(S_i(\theta)\) in
Equation~\ref{eq:rollout-total}. The relation depends only on the grid and the
block sizes, so every head, noise level, and rollout shares it.

The executed mask combines the safety relation with the sampled action. For
query token \(p\in\mathcal Q_q\) and key token \(p'\in\mathcal K_j\) with
\(j\in C_q\), the mask of rollout \(i\) at unit \(u\) is
\begin{equation}
  \mu_{i,u}(p,p')
  = \max\!\Bigl(\mathbf 1\bigl[(p,p')\in\mathcal E_{\mathrm{safe}}\bigr],\,
    a_{i,u,j}\Bigr).
  \label{eq:executed-mask}
\end{equation}
Here \(\mu_{i,u}(p,p')=1\) means that query token \(p\) attends to key token
\(p'\), and \(\mathbf 1[\cdot]\) is the indicator function. Pairs with
\(j\in S_q\) are always attended. The sampled entry \(a_{i,u,j}\) therefore
decides only the unsafe pairs of a candidate block. A candidate that the safety
relation covers in part remains a meaningful action, because dropping it
removes only its unsafe pairs while the safe part stays attended. Every query
token keeps at least its own safety block, so the softmax stays defined even
when a draw keeps no candidate. Intuitively, the route action decides which
non-local context each query block sees, while its local neighborhood stays
deterministic.

\paragraph{Symmetric neighborhood in time.}
Because attention is bidirectional, the neighborhood is symmetric in time as
well as in space, and so is \(\mathcal E_{\mathrm{safe}}\). A query block in
frame \(f\) keeps safe context in frames \(f-1\), \(f\), and \(f+1\), as
Figure~\ref{fig:app-gilbert-safety}(c) shows. No causal constraint removes the
following frame from the neighborhood.

\paragraph{Text conditioning.}
Routing acts only on visual self-attention. Stable Diffusion 3.5 runs a joint
attention over image and text tokens, and we route only the image-to-image
pairs. Every pair that involves a text token stays dense, so each image query
keeps the full text band in addition to its safety neighborhood. The second,
image-only self-attention of Stable Diffusion 3.5 has no text band and relies
on the safety set alone. Wan2.2 injects text through cross-attention, which is
never routed.

\subsection{Block counts in the reference configurations}
\label{app:reference-counts}

\begin{table}[t]
\centering
\caption{\textbf{Token grids, blocks, and candidate sets of the reference
configurations.} Counts
refer to one routed layer and head. Block counts and the safe-pair fraction
cover visual tokens only. The safe-pair fraction is the share of a query
block's visual query-key pairs that lie in \(\mathcal E_{\mathrm{safe}}\).
Means and ranges are taken over query blocks.}
\label{tab:reference-counts}
\small
\begin{tabular}{lcc}
\toprule
 & Wan2.2 & Stable Diffusion 3.5 \\
\midrule
Output size & \(13\times704\times1280\) & \(512\times512\) \\
Token grid \(F\times H\times W\) & \(4\times22\times40\) & \(1\times32\times32\) \\
Visual tokens \(M\) & 3{,}520 & 1{,}024 \\
Block sizes \(B_{\mathrm q}\,/\,B_{\mathrm k}\,/\,B_{\mathrm s}\) & 64 / 16 / 32 & 128 / 32 / 32 \\
Query blocks \(Q\) / key blocks \(J\) & 55 / 220 & 8 / 32 \\
Safety set \(|S_q|\), mean & 18.8 & 1 \\
Candidate set \(|C_q|\), mean [range] & 201.2 [184, 210] & 31 [31, 31] \\
Candidates covered in part, mean & 19.3 & 10.5 \\
Safe-pair fraction, mean [range] & 12.9\% [6.8\%, 21.4\%] & 20.5\% [17.2\%, 23.4\%] \\
Nominal budget \(K_u\) at \(\kappa=0.5\), mean & 100.6 & 16 \\
\bottomrule
\end{tabular}
\end{table}

Table~\ref{tab:reference-counts} reports the token grid, the block counts, and
the candidate-set sizes of both reference configurations. For Wan2.2, the video autoencoder compresses 13 frames at
\(704\times1280\) by a factor of 4 in time and 16 in space, and patchification
halves each spatial side, which gives a \(4\times22\times40\) token grid. For
Stable Diffusion 3.5, a \(512\times512\) image gives a \(32\times32\) image
grid. The text tokens are never routed, so the counts exclude them. In Wan2.2, the
safety relation covers about one eighth of each query block's visual pairs and
leaves about 200 candidate key blocks. In Stable Diffusion 3.5, each 128-token
query block spans four safety blocks, so only one key block is covered in full,
while about a third of the candidates are covered in part. In both generators,
roughly four fifths or more of the visual query-key pairs remain under the
routing policy. The safety set thus guarantees valid local attention while
leaving most of the attention pattern to the learned route.

\section{Implementation details}
\label{app:implementation}
\label{app:extra}

This appendix lists the settings behind Section~\ref{subsec:setup}. The
routing policy and objective are identical for both generators. Only the
routing hyperparameters and optimization settings differ. \rev{Table~\ref{tab:configs}
summarizes both configurations, and Algorithm~\ref{alg:rpo} gives one
training step.}

\paragraph{Stable Diffusion 3.5.}
The reference configuration routes denoising slot 2 with temperature \(T=1\),
fixed offset \(\beta_0=1\), and nominal keep ratio \(\kappa=0.5\). Each keep
probability is therefore \(p_{i,u,j}=\sigma(z_{i,u,j}+1)\), and the observed
keep ratio stays near \(0.58\). Per-head routing pools 128 query and 32 key
tokens per block and builds the safety set from 32-token blocks. It covers
half of the 24 transformer blocks plus the final block, and it routes both the
joint attention and the second image-only self-attention. Each wave contains 64 prompt groups at \(G=2\), split into two
optimizer minibatches of 32 groups. We train rank-32 LoRA adapters with
\(\alpha=64\), learning rate \(5\times10^{-5}\), weight decay \(10^{-4}\), and
no warmup. Advantage processing uses 5\% winsorization, batch-IQR scaling with
floor \(0.03\), and a near-tie gate of \(0.01\). Training runs for 80 steps on
one node with eight H100 GPUs, with evaluation every 5 steps.

\paragraph{Wan2.2.}
The reference configuration uses temperature \(T=2\) and fixed offset
\(\beta_0=-1\), so each keep probability is
\(p_{i,u,j}=\sigma(z_{i,u,j}/2-1)\). We selected this offset as the initial
sparsity operating point. No row-wise root is solved, and the expected number
of retained blocks is not constrained to equal \(K_u\). Routing pools 64 query
tokens and 16 key tokens per block, routes each head independently, and
builds the safety set from 32-token blocks (Appendix~\ref{app:gilbert-safety}). The
configured ratio
\(\kappa=0.5\) sets the nominal normalizer in
Equation~\ref{eq:nominal-budget}, while the realized keep count is recorded
separately. We optimize the trainable generator parameters with learning rate
\(2\times10^{-5}\). Each wave contains 64 prompt groups at \(G=2\) and is
divided into two disjoint optimizer minibatches of 32 groups. The first
minibatch is fresh, and the second contains different pairs from the same
wave. An advantage-spread gate with threshold
\(\tau_{\mathrm{spread}}=4\times10^{-3}\) zeros the advantage of near-tie
pairs, so those pairs contribute no update.
\rev{Each rollout uses the 10-step sampler with classifier-free guidance
scale 5, and the conditional and unconditional branches draw independent
routes. Each rollout routes one of the first five sampler steps. The routed
step follows a schedule over training. Its distribution is concentrated on
slot 4 at step 4, slot 3 at step 20, and slot 2 from step 50 onward, with
linear blending in between. Routing covers half of the 30 transformer blocks
plus the final block. Training updates all generator parameters, warms up the
learning rate over 8 optimizer steps, and runs for 50 steps on two nodes with
eight H100 GPUs each.}

\paragraph{Optimizer safeguards.}
Three generic safeguards stabilize optimization. They are not components of
\RPO{}. First, the second minibatch uses standard PPO ratio clipping with width
\(10^{-4}\) for both generators. Let \(S_i^{\mathrm{old}}\) be the behavior
log-likelihood stored during collection. We use the normalized ratio
\(\rho_i(\theta)=\exp((S_i(\theta)-S_i^{\mathrm{old}})/N_i)\), which leaves
the Bernoulli likelihood of Section~\ref{subsec:conditional-replay} unchanged.
Second, \rev{both generators cap} the processed advantage at
\(A_{\max}^{\mathrm{loss}}=5\). Third, \rev{both generators apply} a global gradient-norm cap\rev{,}
\(\Gamma_{\mathrm{grad}}=1\). \rev{The cap} treats every trainable gradient as one
concatenated vector and rescales its accumulated norm before AdamW consumes
it. \rev{In the reported Wan2.2 run, the largest gradient norm over 50 steps is
\(0.46\), so the cap never fires. Appendix~\ref{app:ablations} ablates the
first safeguard. Removing the second minibatch, and with it ratio clipping,
makes training diverge, although this change also halves the number of
optimizer steps.}

\begin{revblock}
\begin{table}[ht]
\centering
\caption{\textbf{Reference configurations.} Values for the checkpoints
reported in Figure~\ref{fig:sd35-drawbench} (SD3.5) and in
Figures~\ref{fig:dancegrpo-compare} to~\ref{fig:qualitative} and
Table~\ref{tab:vbench} (Wan2.2).}
\label{tab:configs}
\scriptsize
\setlength{\tabcolsep}{4pt}
\begin{tabular}{@{}lll@{}}
\toprule
Setting & SD3.5 Medium & Wan2.2-TI2V-5B \\
\midrule
Rollout & \(512^2\) image, 10 steps, no CFG & 13 frames, \(704\times1280\), 10 steps, CFG 5 \\
Group, prompts per step & \(G=2\), 64 & \(G=2\), 64 \\
Route logit & pooled scaled inner product & pooled scaled inner product \\
Keep probability & \(\sigma(z/T+\beta_0)\), \(T=1\), \(\beta_0=1\) & \(\sigma(z/T+\beta_0)\), \(T=2\), \(\beta_0=-1\) \\
Offset & fixed & fixed \\
Nominal keep ratio \(\kappa\) & 0.5 & 0.5 \\
Routed steps per rollout & 1 (slot 2) & 1 (scheduled slot 4 \(\to\) 3 \(\to\) 2) \\
Routed blocks & 12 of 24 plus final & 15 of 30 plus final \\
Query / key / safety block & 128 / 32 / 32 tokens & 64 / 16 / 32 tokens \\
Trainable parameters & LoRA, rank 32, \(\alpha=64\) & all \\
Learning rate, warmup & \(5\times10^{-5}\), none & \(2\times10^{-5}\), 8 steps \\
Minibatches per step & 2 \(\times\) 32 groups & 2 \(\times\) 32 groups \\
Winsorization, IQR floor & 5\%, 0.03 & 5\%, 0.02 \\
Near-tie gate & 0.01 & 0.004 \\
PPO clip, advantage cap, grad cap & \(10^{-4}\), 5, 1 & \(10^{-4}\), 5, 1 \\
Training steps, GPUs & 80, \(8\times\)H100 & 50, \(16\times\)H100 \\
\bottomrule
\end{tabular}
\end{table}

\begin{algorithm}[ht]
\caption{One \RPO{} training step}
\label{alg:rpo}
\small
\begin{algorithmic}[1]
\Require prompts \(\{c_g\}_{g=1}^{64}\), generator \(\theta\), reward \(r\)
\For{each prompt group \(g\)} \Comment{collection, no gradient}
  \State draw one content seed and one routed level \(t^\star\) shared by both members
  \For{member \(i\in\{1,2\}\)} \Comment{independent routing seeds}
    \State run the deterministic sampler with dense attention except at \(t^\star\)
    \State at \(t^\star\), sample \(a_{i,u,j}\sim\mathrm{Bernoulli}(p_{i,u,j}(\theta))\) unit by unit in forward order
    \State store the mask history \(a_i\), the latent \(x_i^{t^\star}\), and \(S_i^{\mathrm{old}}\)
    \State decode and score \(r_i\)
  \EndFor
  \State \(A^{\mathrm{pair}}_{i}\gets r_i-r_{\bar i}\) \Comment{Equation~\ref{eq:pair-advantage}}
\EndFor
\State \(\widehat A_i\gets\) winsorize, IQR-scale, gate near ties, stop gradient \Comment{Equation~\ref{eq:advantage-pipeline}}
\For{each of the two minibatches of 32 groups}
  \State restart from \(x_i^{t^\star}\), execute the stored masks \(a_i\) in order, and recompute \(S_i(\theta)\) \Comment{Equation~\ref{eq:rollout-total}}
  \State \(\mathcal L\gets\) mean over pairs of \(\mathcal L^{\mathrm{RPO}}\), with PPO ratio clipping on the second minibatch \Comment{Equation~\ref{eq:pg-loss}}
  \State cap the gradient norm and take one AdamW step on \(\theta\)
\EndFor
\end{algorithmic}
\end{algorithm}
\end{revblock}

\section{Robust advantage processing}
\label{app:advantage}

Section~\ref{subsec:paired-objective} converts raw pair differences into a
fixed replay weight. For each reward dimension \(d\), we apply symmetric
winsorization and an interquartile-range scale:
\begin{equation}
\begin{aligned}
  \widetilde A_{i,d}
  &= \clip\!\left(A^{\mathrm{pair}}_{i,d},L_d,U_d\right),
  \qquad
  \mathrm{Scale}_d
  = \max\!\left\{
     \frac{\operatorname{Quantile}_{0.75}(\widetilde A_{\cdot,d})
          -\operatorname{Quantile}_{0.25}(\widetilde A_{\cdot,d})}{1.349},
     \epsilon_{\mathrm s}
     \right\},
  \\
  \widehat A_i
  &= \sg\!\left(
     \sum_{d\in\mathcal D}
     w_d\frac{\widetilde A_{i,d}}{\mathrm{Scale}_d}
     \right).
\end{aligned}
  \label{eq:advantage-pipeline}
\end{equation}
Here \(L_d\) and \(U_d\) are the symmetric winsorization thresholds,
\(\operatorname{Quantile}_{\alpha}\) is the empirical quantile at probability
\(\alpha\), \(\epsilon_{\mathrm s}>0\) is a numerical floor, and \(w_d\) weights
reward dimension \(d\in\mathcal D\). The constant \(1.349\) converts the
interquartile range into a standard-deviation scale under a Gaussian
reference. The stop-gradient operator \(\sg(\cdot)\) makes \(\widehat A_i\) a
fixed replay weight. Because the thresholds are symmetric and the processing
is shared across both signs of every pair, it preserves pair antisymmetry up
to numerical precision. Our experiments use a single HPSv3 dimension, so
\(\mathcal D\) contains one element.

\section{Evaluation protocols}
\label{app:protocol}

\paragraph{Wan2.2 dense probe.}
The reward scores the rendered video rather than the weights. The same
pretraining checkpoint can therefore produce different step-zero scores under
different rollout \rev{and evaluation paths}. On the 61-prompt probe, the \RPO{} \rev{run} starts
at 4.92 raw HPSv3, \rev{the DanceGRPO run at 5.53, and the Flow-GRPO run at
4.71. The runs differ in rollout backend, numerical precision, and validation
seeding, so the gap is not a single renderer offset.}
Figure~\ref{fig:dancegrpo-eval} therefore plots every dense-model number as a
gain over the run's own step-zero score. \rev{This removes the step-zero
difference but cannot rule out effects that change during training. The
VBench evaluation (Table~\ref{tab:vbench} and Table~\ref{tab:vbench-full})
avoids this issue. It renders every Wan2.2 checkpoint with one shared
pipeline, prompt set, and set of initial noises, and it reports absolute
scores.}

\paragraph{Comparison with DanceGRPO.}
The DanceGRPO run uses the same generator, reward, and probe prompts as \RPO{}.
Four differences remain. First, DanceGRPO is full-model GRPO with \(G=8\) and
a stochastic SDE sampler, whereas \RPO{} routes a single denoising transition
with \(G=2\). Second, the two runs use different rollout backends. Third,
the two runs record dense evaluation through different evaluation code,
although both measure raw HPSv3 under dense inference on the same prompts.
Fourth, the DanceGRPO run follows a longer 120-step schedule, so
Figure~\ref{fig:dancegrpo-compare} shows only its early phase. \rev{The
Flow-GRPO run also follows a 120-step schedule. Table~\ref{tab:schedules}
reports both full runs.}

\begin{revblock}
\paragraph{Full training schedules.}
Table~\ref{tab:schedules} lists the dense-probe gain of each run over its
own step-zero score, together with the number of rollout videos generated.
Because each gain is measured against the run's own step zero, these numbers
compare learning progress within runs. \RPO{} reaches \(+2.96\) at step 15
after \(1{,}920\) videos and stays between \(+2.86\) and \(+2.98\) at every
probe from step 15 to step 50. Flow-GRPO first exceeds this
level between steps 50 and 60, after \(25{,}600\) to \(30{,}720\) videos,
and then keeps raising its dense-probe gain. DanceGRPO peaks at \(+0.32\) at
step 55, falls to \(-2.32\) at step 85, and ends its full run at \(-0.04\).
Table~\ref{tab:shared-renderer} repeats this comparison with absolute scores
from one shared renderer.
Flow-GRPO saved checkpoints only at steps 40, 80, and 120. Its step-40
checkpoint has a dense gain of \(+2.04\), yet it falls below the base model
on 11 of the 16 VBench dimensions and on the VBench total
(Table~\ref{tab:vbench-full}).

\begin{table}[ht]
\centering
\caption{\textbf{Dense-probe HPSv3 gain over each run's own step-zero score on
Wan2.2.} A dash marks a step that was not evaluated or lies beyond the run's
schedule. Rollout videos per step are 128 for \RPO{} and 512 for both
baselines.}
\label{tab:schedules}
\scriptsize
\setlength{\tabcolsep}{3.2pt}
\begin{tabular}{@{}lrrrrrrrrrr@{}}
\toprule
Step & 5 & 10 & 15 & 20 & 30 & 40 & 50 & 60 & 80 & 100 \\
\midrule
\RPO{} & \(+2.17\) & \(+2.78\) & \(+2.96\) & \(+2.97\) & \rev{\(+2.94\)} & \rev{\(+2.93\)} & \(+2.86\) & -- & -- & -- \\
DanceGRPO & \(+0.29\) & \rev{\(+0.28\)} & \rev{\(+0.25\)} & \rev{\(+0.15\)} & \rev{\(+0.01\)} & \rev{\(+0.14\)} & \(+0.17\) & \rev{\(-0.02\)} & \rev{\(-1.39\)} & \rev{\(-1.16\)} \\
Flow-GRPO & \(-0.08\) & \(-3.05\) & \(-1.78\) & \(+0.87\) & \(+1.41\) & \(+2.04\) & \(+2.69\) & \(+3.18\) & \(+3.80\) & \(+4.56\) \\
\midrule
Videos (k), \RPO{} & 0.6 & 1.3 & 1.9 & 2.6 & 3.8 & 5.1 & 6.4 & -- & -- & -- \\
Videos (k), baselines & 2.6 & 5.1 & 7.7 & 10.2 & 15.4 & 20.5 & 25.6 & 30.7 & 41.0 & 51.2 \\
\bottomrule
\end{tabular}
\end{table}

\paragraph{Training cost accounting.}
\label{app:cost}
Table~\ref{tab:cost} compares the training cost of the three Wan2.2 runs. All
runs use 64 prompts per step, the same model, 13 frames at
\(704\times1280\), and 10 denoising steps. Every cost is measured from the
training logs and includes startup, validation every 5 steps, and checkpoint
saves. The baselines roll out with the vLLM-Omni backend. The routed step of
\RPO{} needs FlexAttention, which vLLM-Omni does not support, so \RPO{} rolls
out with a batched diffusers backend instead. A full 50-step run of the
reference configuration takes \(25.4\) GPU-hours. The saving has two
sources. Each step generates 4\(\times\) fewer rollout videos, and
each video replays one transition rather than the two-step SDE window of the
baselines. VBench evaluation is excluded from every cost.

\begin{table}[ht]
\centering
\caption{\textbf{Training budget and measured training cost of the Wan2.2
runs.} GPU-hours are measured from the training logs and include startup,
validation, and checkpoint saves.}
\label{tab:cost}
\scriptsize
\setlength{\tabcolsep}{4pt}
\begin{tabular}{@{}lrrr@{}}
\toprule
 & DanceGRPO & Flow-GRPO & \RPO{} \\
\midrule
Rollout videos per step & 512 & 512 & 128 \\
Replayed transitions per video & 2 & 2 & 1 \\
Rollout videos to step 50 & 25{,}600 & 25{,}600 & 6{,}400 \\
Rollout backend & vLLM-Omni & vLLM-Omni & diffusers, batched \\
\midrule
GPU-hours to step 50 & 102.3 & 93.1 & 25.4 \\
GPU-hours to the evaluated checkpoint & 102.3 (step 50) & 75.4 (step 40) & 25.4 (step 50) \\
\bottomrule
\end{tabular}
\end{table}
\end{revblock}

\paragraph{VBench.}
We evaluate all 16 VBench dimensions on the standard prompt suite. Each Wan2.2
model generates \(6{,}220\) videos with 13 frames at \(704\times1280\) and
24 fps, using a deterministic 50-step ODE sampler with classifier-free guidance
scale 5.0. The four Wan2.2 models share prompts and initial noise.
CogVideoX-5B and HunyuanVideo use the same prompts, frame count, denoising
steps, and scoring, but they keep their native resolution and guidance. Their
rows therefore provide a reference under matched content and scoring, although
their generation settings and training compute differ. Training GPU-hours use
a common accounting scope and exclude VBench evaluation\rev{
(Appendix~\ref{app:cost})}.

\begin{revblock}
Table~\ref{tab:vbench-full} reports all 16 dimensions, the quality and
semantic scores, and the total. It adds the \RPO{} checkpoint at step 20. The
rate-type dimensions average 360 to 500 videos per model, so one model's
standard error on these dimensions is about 2 points. The Wan2.2 models share
prompts and initial noise, which makes their differences tighter than this.
Relative to the base model, \RPO{} improves 10 of the 16 dimensions. It
lowers background consistency, temporal flickering, color, appearance style,
temporal style, and overall consistency. The largest loss is appearance style
at \(-3.14\).

\begin{table}[ht]
\centering
\caption{\textbf{All 16 VBench dimensions on Wan2.2-TI2V-5B (\%).} The Wan2.2
columns share prompts, initial noise, and the generation configuration.
CogVideoX-5B and HunyuanVideo use the same prompts, frame count, steps, and
scoring at their native resolution and guidance.}
\label{tab:vbench-full}
\scriptsize
\setlength{\tabcolsep}{3.5pt}
\begin{tabular}{@{}lrrrrrrr@{}}
\toprule
Dimension & Base & \RPO{}@20 & \RPO{}@50 & DanceGRPO@50 & Flow-GRPO@40 & CogVideoX-5B & HunyuanVideo \\
\midrule
Subject consistency & 98.37 & 98.70 & 98.69 & 98.38 & 97.59 & 96.50 & 98.76 \\
Background consistency & 98.50 & 98.23 & 98.22 & 98.32 & 97.77 & 96.99 & 99.10 \\
Temporal flickering & 99.38 & 99.10 & 99.10 & 99.38 & 98.91 & 98.32 & 99.75 \\
Motion smoothness & 98.76 & 99.04 & 99.06 & 98.75 & 97.88 & 97.42 & 99.42 \\
Dynamic degree & 50.56 & 51.67 & 52.22 & 50.28 & 63.06 & 89.17 & 28.06 \\
Aesthetic quality & 61.91 & 62.62 & 62.64 & 61.74 & 60.09 & 60.17 & 63.05 \\
Imaging quality & 67.71 & 68.91 & 68.65 & 67.69 & 67.43 & 61.92 & 63.78 \\
Object class & 81.01 & 84.49 & 85.41 & 80.81 & 79.10 & 82.44 & 65.79 \\
Multiple objects & 59.15 & 63.37 & 61.30 & 59.34 & 59.98 & 57.29 & 56.46 \\
Human action & 79.40 & 84.20 & 84.20 & 80.00 & 81.80 & 86.40 & 83.20 \\
Color & 92.51 & 89.88 & 91.16 & 90.77 & 89.43 & 85.02 & 84.88 \\
Spatial relationship & 74.79 & 75.83 & 75.18 & 73.30 & 75.24 & 53.96 & 57.16 \\
Scene & 37.41 & 41.47 & 40.84 & 36.95 & 36.73 & 42.73 & 27.01 \\
Appearance style & 23.15 & 20.08 & 20.02 & 23.08 & 21.52 & 23.47 & 20.13 \\
Temporal style & 24.34 & 24.32 & 24.28 & 24.28 & 24.58 & 23.59 & 21.56 \\
Overall consistency & 25.06 & 24.68 & 24.67 & 25.04 & 24.95 & 25.64 & 21.96 \\
\midrule
Quality score & 83.98 & 84.40 & 84.41 & 83.90 & 83.67 & 84.00 & 82.52 \\
Semantic score & 72.12 & 72.56 & 72.38 & 71.71 & 71.29 & 70.50 & 63.37 \\
Total score & 81.61 & 82.03 & 82.00 & 81.46 & 81.19 & 81.30 & 78.69 \\
\bottomrule
\end{tabular}
\end{table}
\end{revblock}

\paragraph{Stable Diffusion 3.5.}
During training, dense evaluation uses a 28-step sampler and reports raw
HPSv3 on 64 held-out prompts. Early runs draw this subset with the training
seed. For controlled seed comparisons, we instead use a fixed
64-prompt file and a fixed validation noise seed. The DrawBench evaluation in
Figure~\ref{fig:sd35-drawbench} uses \(1{,}000\) images from \(200\) prompts
and \(5\) seeds, generated without classifier-free guidance\rev{ and with 40
sampling steps. Image \(k\) uses the same initial noise for the base and
trained models, and every metric difference in the figure is at least four
standard errors over the 200 prompts}.

\section{Extended discussion}
\label{app:discussion}

\paragraph{Known failure modes.}
Two failure modes remain open. The first is a dense-gain fade under certain
routing schedules. In these runs, the dense-inference gain peaks early and then
decays over later training steps. \rev{The reported Wan2.2 run does not
show this fade within its budget. Its dense gain stays between \(+2.86\) and
\(+2.98\) at every probe from step 15 to step 50 (Table~\ref{tab:schedules}),
and its VBench totals at steps 20 and 50 agree (Table~\ref{tab:vbench-full}).
In the sweep of Appendix~\ref{app:ablations},
every run with winsorized advantage processing improves, whereas three of
eight runs without it collapse or drift below their initial score. The
mechanism behind the fade is not yet isolated.} The second is reward hacking against the frozen
preference model. This risk is general to online RL for generative models and
is not specific to \RPO{}. Both failure modes suggest that a scalar terminal
reward is a coarse supervisor for a discrete routing action.
\rev{On Wan2.2, both training seeds of the reference configuration improve
the dense model and score above the base model on VBench
(Appendix~\ref{app:ablations}).}

\paragraph{Estimator scope.}
The recorded-action likelihood in Equation~\ref{eq:rollout-total} is exact,
but the resulting update is a route-channel semi-gradient. It omits the
pathwise gradient of the generator through later denoising transitions and
through the frozen reward model. The fixed offset also sets only an initial
sparsity operating point. It does not preserve the expected keep count as the
attention affinities evolve, so reward improvement can coincide with a change
in mask density. \rev{In the reference Wan2.2 run the density stays stable.
The realized keep ratio is \(0.48\) at step 1, \(0.48\) at step 15, and
\(0.47\) at step 50, against the nominal \(\kappa=0.5\). Among the
runs for which we recorded the keep ratio, only the diverging run of
Table~\ref{tab:components} shows a large density change. Its keep ratio falls
to \(0.41\) and then rises to about \(0.60\).}

\paragraph{Extensions.}
Two directions extend the method naturally. Multi-step routing would route
more than one denoising transition per rollout. It needs a coupled-pair
estimator to keep the matched-pair advantage well defined. Larger groups would
give a leave-one-out baseline with more members. They trade part of the
pair's variance cancellation for a more stable advantage scale. Both
extensions are compatible with the estimator as stated.

\begin{revblock}
\section{Additional ablations and statistics}
\label{app:ablations}

\paragraph{Advantage processing, routed-step sampling, and seeds.}
Table~\ref{tab:sweep} varies the safeguards of Appendix~\ref{app:advantage}
and the routed-step sampler on Wan2.2 at the reference setting: 13 frames,
50 training steps, and two training seeds per arm. The fixed sampler routes
slot 3 for every rollout. The i.i.d.\ sampler draws the routed slot per rollout
from a distribution that moves from slot 4 toward earlier slots during
training. The reference run adds the slot schedule of
Appendix~\ref{app:implementation}. With winsorization, all eight runs improve
the dense model, by \(+1.06\) to \(+2.86\) with a mean of \(+2.29\), and the
near-tie gate and the sampler change the gain only moderately. Without
winsorization, the same changes swing runs between collapse and \(+2.86\), and
three of eight runs collapse or drift below their initial score. Both seeds of
the reference configuration improve, by \(+2.86\) and \(+2.40\). We selected
the reference configuration from this sweep. The sampler-space baselines in
Section~\ref{sec:experiments} are single runs.

\begin{table}[ht]
\centering
\caption{\textbf{Dense-probe HPSv3 gain at step 50 on Wan2.2 for two training
seeds.} Collapse marks a run whose dense score fell sharply near step 25.}
\label{tab:sweep}
\scriptsize
\setlength{\tabcolsep}{4pt}
\begin{tabular}{@{}lllcrr@{}}
\toprule
Arm & Routed-step sampler & Near-tie gate & Winsorization & Seed 42 & Seed 43 \\
\midrule
S1 & fixed (slot 3) & 0.002 & on & \(+2.68\) & \(+1.06\) \\
S2 & fixed (slot 3) & 0.004 & on & \(+2.72\) & \(+2.07\) \\
S3 & fixed (slot 3) & 0.002 & off & \(+2.86\) & \(+1.60\) \\
S4 & fixed (slot 3) & 0.004 & off & \(-1.41\) (collapse) & \(-0.25\) \\
S5 & i.i.d.\ & 0.004 & off & \(+1.56\) & \(+1.88\) \\
S6 & i.i.d.\ & 0.006 & off & \(-0.09\) (collapse) & \(+2.31\) \\
S7 & i.i.d.\ & 0.004 & on & \(+2.24\) & \(+2.25\) \\
Reference & i.i.d., slot schedule & 0.004 & on & \(+2.86\) & \(+2.40\) \\
\bottomrule
\end{tabular}
\end{table}

\paragraph{Why matched pairs.}
A diagnostic probe measures how strongly the reward responds to the route
compared with the content seed. It uses the base Wan2.2 model at a shorter
5-frame setting with 8 prompts and 16 pairs per cell. Two routes under the
same content seed differ by \(|\Delta r|=0.26\) raw HPSv3 at a realized keep
ratio of 0.49. In contrast, the same route under two content seeds differs by
\(0.96\). Across keep ratios from 0.20 to 0.92, the seed effect is about 3 to 5
times larger than the route effect. An unmatched pair would therefore
measure mostly seed variation, whereas the matched pair isolates the
routing action. This probe measures reward differences, not gradient
variance. The component ablations below test whether this attribution also
changes training.

\begin{table}[ht]
\centering
\caption{\textbf{Reward sensitivity of the base Wan2.2 model to the route and
to the content seed} (raw HPSv3, 5 frames, 16 pairs per cell).}
\label{tab:pair-probe}
\scriptsize
\begin{tabular}{@{}lrrrr@{}}
\toprule
Realized keep ratio & 0.20 & 0.49 & 0.79 & 0.92 \\
\midrule
Same seed, two routes, \(|\Delta r|\) & 0.29 & 0.26 & 0.36 & 0.17 \\
Same route, two seeds, \(|\Delta r|\) & 1.05 & 0.96 & 1.02 & 0.87 \\
Ratio & 0.28 & 0.27 & 0.36 & 0.20 \\
\bottomrule
\end{tabular}
\end{table}

\paragraph{Shared dense renderer.}
The dense-probe scores of Figure~\ref{fig:dancegrpo-eval} come from each
run's own validation path, so their step-zero scores differ
(Appendix~\ref{app:protocol}). Table~\ref{tab:shared-renderer} removes this
difference. One renderer and one HPSv3 scorer render saved checkpoints of
the three runs on the 61 probe prompts, with the 10-step sampler, guidance
scale 5, and the same prompt-keyed initial noise for every model. All runs
start from the same pretrained weights, so they share one step-zero score of
\(5.29\), and every entry is an absolute score. \RPO{} reaches \(7.87\) at
step 20 and \(7.82\) at step 50. Flow-GRPO reaches \(6.63\) at step 40, after
8 times the rollout videos that \RPO{} used by step 20, and first exceeds the
\RPO{} level at step 80, after 16 times. DanceGRPO stays at the base level.
The comparison therefore supports a smaller rollout budget for a given reward,
not a higher final reward. The \RPO{} checkpoints at steps 10, 30, and 40 and
the DanceGRPO checkpoint at step 120 are not part of this evaluation.

\begin{table}[ht]
\centering
\caption{\textbf{Absolute dense-probe HPSv3 of saved Wan2.2 checkpoints under
one shared renderer} (61 prompts, identical initial noise). Prompts improved
counts probe prompts that score above the base model.}
\label{tab:shared-renderer}
\scriptsize
\setlength{\tabcolsep}{4pt}
\begin{tabular}{@{}lrrrr@{}}
\toprule
Checkpoint & Step & Rollout videos & Absolute HPSv3 & Prompts improved \\
\midrule
Base model & 0 & 0 & 5.29 & -- \\
\RPO{} & 20 & 2{,}560 & \textbf{7.87} & 59/61 \\
\RPO{} & 50 & 6{,}400 & 7.82 & 57/61 \\
DanceGRPO & 50 & 25{,}600 & 5.32 & 33/61 \\
Flow-GRPO & 40 & 20{,}480 & 6.63 & 43/61 \\
Flow-GRPO & 80 & 40{,}960 & 8.42 & 60/61 \\
\bottomrule
\end{tabular}
\end{table}

\paragraph{VBench seeds, later baseline checkpoints, and uncertainty.}
Table~\ref{tab:vbench-extra} adds three checkpoints to the VBench evaluation
under the protocol of Table~\ref{tab:vbench-full}. The second training seed of
the reference configuration scores \(82.24\), so both seeds exceed the base
model (\(81.61\)). The two seeds improve different dimensions. The largest
per-dimension gains of seed 42 are semantic (object class, human action, and
scene), although losses in appearance style and color keep its semantic score
at \(+0.26\). Seed 43 raises dynamic degree to \(69.72\) and the quality score
to \(84.90\) but stays slightly below the base model on the semantic score. The seed-43 checkpoint in Table~\ref{tab:vbench-extra} comes from a separate
training run with the same configuration and seed. This run was resumed once
from its step-40 checkpoint, which restarts the optimizer state. The seed-43
dense gain of \(+2.40\) in Table~\ref{tab:sweep} refers to the earlier run with
this seed. Flow-GRPO at steps 80 and 120
matches the \RPO{} total (\(82.02\) and \(81.99\)) after \(6.4\) and \(9.6\)
times the rollout videos of \RPO{} at step 50. Its semantic score keeps rising
to \(75.26\), whereas its dynamic degree falls to \(33.89\), below the base
model.

Table~\ref{tab:bootstrap} quantifies the uncertainty of the seed-42
comparison with a paired bootstrap. We resample prompts with replacement,
apply the same resample to both models, and recompute the VBench quality,
semantic, and total scores with VBench's own normalization and weights over
\(2{,}000\) resamples. The quality-score gain of \RPO{}, \(+0.43\), has a 95\%
interval of \([+0.02,+0.85]\) that excludes zero. The total gain,
\(+0.39\) \([-0.01,+0.79]\), narrowly includes zero, so we read it as
directional. Four dimensions improve and three decline with intervals that
exclude zero. Against the same base model, the total changes by \(-0.15\)
\([-0.35,+0.04]\) for DanceGRPO at step 50 and by \(-0.42\) \([-0.81,-0.04]\)
for Flow-GRPO at step 40.

\begin{table}[ht]
\centering
\caption{\textbf{VBench for the second \RPO{} training seed and later
Flow-GRPO checkpoints (\%).} Same prompts, initial noise, and generation
configuration as Table~\ref{tab:vbench-full}. Rollout videos to each checkpoint
are \(6{,}400\) for \RPO{} and \(20{,}480\), \(40{,}960\), and \(61{,}440\) for
Flow-GRPO at steps 40, 80, and 120.}
\label{tab:vbench-extra}
\scriptsize
\setlength{\tabcolsep}{3.5pt}
\begin{tabular}{@{}lrrrrrr@{}}
\toprule
Dimension & Base & \RPO{}@50, seed 42 & \RPO{}@50, seed 43 & Flow-GRPO@40 & Flow-GRPO@80 & Flow-GRPO@120 \\
\midrule
Subject consistency & 98.37 & 98.69 & 97.96 & 97.59 & 98.30 & 98.80 \\
Background consistency & 98.50 & 98.22 & 97.71 & 97.77 & 98.36 & 98.62 \\
Temporal flickering & 99.38 & 99.10 & 99.18 & 98.91 & 99.10 & 99.19 \\
Motion smoothness & 98.76 & 99.06 & 98.81 & 97.88 & 98.23 & 98.72 \\
Dynamic degree & 50.56 & 52.22 & 69.72 & 63.06 & 51.94 & 33.89 \\
Aesthetic quality & 61.91 & 62.64 & 61.32 & 60.09 & 62.75 & 64.51 \\
Imaging quality & 67.71 & 68.65 & 66.58 & 67.43 & 70.02 & 71.41 \\
Object class & 81.01 & 85.41 & 82.20 & 79.10 & 81.98 & 84.73 \\
Multiple objects & 59.15 & 61.30 & 62.29 & 59.98 & 68.69 & 73.69 \\
Human action & 79.40 & 84.20 & 82.60 & 81.80 & 82.20 & 85.60 \\
Color & 92.51 & 91.16 & 91.77 & 89.43 & 89.74 & 89.51 \\
Spatial relationship & 74.79 & 75.18 & 74.96 & 75.24 & 74.32 & 78.90 \\
Scene & 37.41 & 40.84 & 36.98 & 36.73 & 41.05 & 39.53 \\
Appearance style & 23.15 & 20.02 & 20.33 & 21.52 & 21.93 & 22.41 \\
Temporal style & 24.34 & 24.28 & 24.29 & 24.58 & 24.85 & 24.81 \\
Overall consistency & 25.06 & 24.67 & 24.66 & 24.95 & 25.50 & 25.57 \\
\midrule
Quality score & 83.98 & 84.41 & 84.90 & 83.67 & 84.13 & 83.67 \\
Semantic score & 72.12 & 72.38 & 71.60 & 71.29 & 73.55 & 75.26 \\
Total score & 81.61 & 82.00 & 82.24 & 81.19 & 82.02 & 81.99 \\
\bottomrule
\end{tabular}
\end{table}

\begin{table}[ht]
\centering
\caption{\textbf{Paired bootstrap of the VBench difference between \RPO{}@50
(seed 42) and the base model} (percentage points, 2{,}000 resamples over
prompts, 95\% intervals). $^\ast$The interval excludes zero.}
\label{tab:bootstrap}
\scriptsize
\setlength{\tabcolsep}{4pt}
\begin{tabular}{@{}lrr@{\hspace{14pt}}lrr@{}}
\toprule
Dimension & Difference & Interval & Dimension & Difference & Interval \\
\midrule
Subject consistency & \(+0.32^\ast\) & \([+0.17,+0.51]\) & Human action & \(+4.80^\ast\) & \([+1.60,+8.20]\) \\
Background consistency & \(-0.28^\ast\) & \([-0.42,-0.14]\) & Color & \(-1.35\) & \([-4.06,+1.15]\) \\
Temporal flickering & \(-0.28^\ast\) & \([-0.37,-0.18]\) & Spatial relationship & \(+0.39\) & \([-3.51,+4.46]\) \\
Motion smoothness & \(+0.30^\ast\) & \([+0.21,+0.38]\) & Scene & \(+3.43\) & \([-1.03,+7.99]\) \\
Dynamic degree & \(+1.67\) & \([-3.06,+6.39]\) & Appearance style & \(-3.14^\ast\) & \([-3.72,-2.62]\) \\
Aesthetic quality & \(+0.72\) & \([-0.01,+1.47]\) & Temporal style & \(-0.06\) & \([-0.42,+0.29]\) \\
Imaging quality & \(+0.94\) & \([0.00,+1.95]\) & Overall consistency & \(-0.39\) & \([-0.80,+0.05]\) \\
Object class & \(+4.40^\ast\) & \([+1.27,+7.44]\) & Quality score & \(+0.43^\ast\) & \([+0.02,+0.85]\) \\
Multiple objects & \(+2.15\) & \([-2.70,+6.83]\) & Semantic score & \(+0.26\) & \([-0.88,+1.40]\) \\
 & & & Total score & \(+0.39\) & \([-0.01,+0.79]\) \\
\bottomrule
\end{tabular}
\end{table}

\paragraph{Component ablations.}
Table~\ref{tab:components} removes one component of the reference
configuration at a time, with seed 42 and all other settings fixed. Rows
marked with an asterisk were trained on a different hardware setup. Permuting
the route logits across the candidate blocks of each
routing unit keeps the set of keep probabilities, and the realized keep ratio
stays at \(0.50\) over the recorded steps. It removes only the affinity pattern.
This variant plateaus at about \(+2.6\) through step 50, roughly \(0.3\) below the reported run at
matched steps. Drawing independent content seeds
for the two members of each pair reaches \(+2.73\) at step 20 and \(+1.88\) at
step 50. This variant also changes the noise of the dense probe, with a
step-zero score of \(4.26\) instead of \(4.92\), so all rows compare gains over
each run's own step zero. Both variants keep a large gain. \RPO{} therefore
learns without matched content seeds, although with single runs we cannot
attribute the step-50 gap of the independent-seed run to matching. Most of
the gain also does not depend on the specific affinity pattern. In contrast, removing the stale second minibatch,
and with it ratio clipping, makes training diverge to \(-13.81\) at step 50.
Its keep ratio first falls to \(0.41\) and later rises to about \(0.60\).
This variant also halves the number of optimizer steps. It therefore shows
that the two-minibatch update is necessary in this configuration, but it does
not isolate the effect of ratio clipping.

The last row is a sampler-space baseline at the rollout budget of \RPO{}.
Flow-GRPO with \(G=2\), shared initial noise, and 64 prompts per step
generates the same 128 rollout videos per step. Its gains are measured on its
own validation path, whose step-zero score is \(5.54\). At step 15, the three
non-diverging \RPO{} runs of Table~\ref{tab:components} gain \(+2.44\) to
\(+2.96\), whereas this baseline gains \(+0.67\). It then falls to
\(-0.42\) at step 35 and ends at \(+1.28\), below the \RPO{} runs at step 50
(\(+1.88\) to \(+2.86\)). At an equal budget, routing therefore learns faster
and more stably.

\begin{table}[ht]
\centering
\caption{\textbf{Component ablations on Wan2.2 (seed 42).} Dense-probe HPSv3
gain over each run's own step zero. $^\ast$Trained on a different hardware setup.}
\label{tab:components}
\scriptsize
\setlength{\tabcolsep}{3.5pt}
\begin{tabular}{@{}llrrrrr@{}}
\toprule
Variant & Removed component & Step 5 & Step 15 & Step 20 & Step 35 & Step 50 \\
\midrule
Reference (reported run) & none & \(+2.17\) & \(+2.96\) & \(+2.97\) & \(+2.87\) & \(+2.86\) \\
Permuted route logits\(^\ast\) & affinity pattern, same density & \(+2.07\) & \(+2.67\) & \(+2.63\) & \(+2.59\) & \(+2.65\) \\
Independent content seeds\(^\ast\) & matched content seed & \(+1.88\) & \(+2.44\) & \(+2.73\) & \(+2.46\) & \(+1.88\) \\
Single fresh minibatch\(^\ast\) & stale update, ratio clipping & \(-1.78\) & \(-7.05\) & \(-7.48\) & \(-13.65\) & \(-13.81\) \\
\midrule
Flow-GRPO, \(G=2\), shared noise\(^\ast\) & routing (SDE exploration instead) & \(+0.81\) & \(+0.67\) & \(+0.57\) & \(-0.42\) & \(+1.28\) \\
\bottomrule
\end{tabular}
\end{table}

\paragraph{Gradient variance of matched and unmatched pairs.}
At the step-20 checkpoint of the reported run, we draw 32 matched pairs and
32 pairs with independent content seeds and compute the gradient of
\(\mathcal L^{\mathrm{RPO}}\) for each pair. The gradient covers the
self-attention query, key, value, and output projections of the last two
transformer blocks (\(75.5\)M parameters) and uses the advantage processing
of the reference configuration. The mean pairwise cosine similarity between
per-pair gradients is \(-0.002\) for matched pairs and \(+0.004\) for
unmatched pairs, so the per-pair gradients are nearly orthogonal under both
protocols. The trace of the gradient covariance is \(5.8\times10^{-6}\) for
matched and \(3.1\times10^{-6}\) for unmatched pairs. This ratio does not show
a variance reduction from matching. It is also confounded by the batch-IQR
scaling of Appendix~\ref{app:advantage}. Matched pairs have a smaller reward
spread, so the scaling enlarges their advantages and gradients, with mean
gradient norms of \(1.9\times10^{-3}\) and \(1.4\times10^{-3}\). We therefore
find no measurable benefit of matched content seeds on gradient variance at
this scale. The matched pair still attributes each reward difference to the
routing action (Table~\ref{tab:pair-probe}), and \RPO{} also learns without it.
\end{revblock}

\end{document}